\documentclass[a4paper]{article}
\usepackage{xcolor}
\usepackage[nottoc,notlot,notlof]{tocbibind}
\definecolor{dandelion}{RGB}{255, 200, 0} 
\definecolor{cornflowerblue}{RGB}{100, 149, 237}
\definecolor{pastelgray}{RGB}{190, 190, 190}
\definecolor{pastelyellow}{RGB}{252, 229, 153}
\definecolor{pastelpurple}{RGB}{215, 189, 226}
\definecolor{pastelred}{RGB}{245, 183, 177}

\usepackage{amssymb} 
\usepackage{pifont}  

\usepackage[english]{babel}
\usepackage{layout}
\usepackage{natbib}
\usepackage{amsmath}
\usepackage{mathrsfs}
\usepackage{amssymb}
\usepackage{changes}
\usepackage{graphicx}
\usepackage{array}
\usepackage{caption} 
\usepackage{adjustbox} 
\usepackage{tcolorbox}

\renewcommand{\arraystretch}{1.5} 
\usepackage{multirow} 

\usepackage{authblk}

\usepackage[euler]{textgreek}
\usepackage{ucs}

\usepackage{ifthen}

\usepackage{float}
\usepackage{lipsum}
\usepackage{listing}
\usepackage{makecell} 

\usepackage{bibentry}
\usepackage{appendix}
\usepackage{upgreek}
\usepackage{chngcntr}
\usepackage{listings}
\usepackage{setspace}
\usepackage{enumitem}
\usepackage{cite}
\usepackage{pgfgantt}
\usepackage[colorlinks=true,citecolor=blue]{hyperref}%
\usepackage{soul}

\usepackage{multirow}   
\usepackage{booktabs}   

\usepackage{arydshln} 
\usepackage{array}   
\newcolumntype{C}[1]{>{\centering\arraybackslash}p{#1}}

\usepackage[a4paper,top=2cm, bottom=2cm, left=2cm, right=2cm]{geometry}

\usepackage{newfloat}
\DeclareFloatingEnvironment[name={Appendix-Figure}]{suppfigure}
\DeclareFloatingEnvironment[name={Appendix-Table}]{supptable}

\usepackage[capitalize,noabbrev]{cleveref}
\crefformat{figure}{#2\figurename~#1#3}
\crefformat{table}{#2\tablename~#1#3}
\crefname{suppfigure}{Appendix 1-Figure}{Appendix 1-Figures}
\crefname{supptable}{Appendix 1-Table}{Appendix 1-Tables}
\crefformat{suppfigure}{#2\suppfigurename~#1#3}
\crefformat{supptable}{#2\supptablename~#1#3}
\crefname{equation}{Equation}{Equations}
\creflabelformat{equation}{#2#1#3}

\newlist{todolist}{itemize}{2}
\setlist[todolist]{label=$\square$}
\usepackage{pifont}

\definecolor{darkgreen}{rgb}{0.0, 0.8, 0.0}

\newcommand{\cmark}{\textcolor{darkgreen}{\ding{51}}} 
\newcommand{\xmark}{\textcolor{red}{\ding{55}}}       

\definecolor{lightblue}{RGB}{221,233,248}
\definecolor{lightgray}{RGB}{230,230,230}

\tcbset{
  humanbubble/.style={
    colback=lightblue!20,
    colframe=lightblue,
    boxrule=0.5mm,
    arc=2mm,
    left=1mm, right=1mm, top=0mm, bottom=0mm, 
    before skip=0pt, after skip=0pt, 
  },
  assistantbubble/.style={
    colback=lightgray!20,
    colframe=lightgray,
    boxrule=0.5mm,
    arc=2mm,
    left=1mm, right=1mm, top=0mm, bottom=0mm, 
    before skip=0pt, after skip=0pt, 
  }
}

\definecolor{lightred}{RGB}{255,220,220}
\definecolor{lightbrown}{RGB}{245,222,179}
\definecolor{lightgreen}{RGB}{210,245,210}
\definecolor{lightpurple}{RGB}{240,230,255}

\tcbset{
  bubbleblue/.style={
    colback=lightblue!20,
    colframe=lightblue,
    boxrule=0.5mm,
    arc=2mm,
    left=1mm, right=1mm, top=0mm, bottom=0mm,
    before skip=0pt, after skip=0pt,
  },
  bubblegray/.style={
    colback=lightgray!20,
    colframe=lightgray,
    boxrule=0.5mm,
    arc=2mm,
    left=1mm, right=1mm, top=0mm, bottom=0mm,
    before skip=0pt, after skip=0pt,
  },
  bubblered/.style={
    colback=lightred!20,
    colframe=lightred,
    boxrule=0.5mm,
    arc=2mm,
    left=1mm, right=1mm, top=0mm, bottom=0mm,
    before skip=0pt, after skip=0pt,
  },
  bubblebrown/.style={
    colback=lightbrown!20,
    colframe=lightbrown,
    boxrule=0.5mm,
    arc=2mm,
    left=1mm, right=1mm, top=0mm, bottom=0mm,
    before skip=0pt, after skip=0pt,
  },
  bubblegreen/.style={
    colback=lightgreen!20,
    colframe=lightgreen,
    boxrule=0.5mm,
    arc=2mm,
    left=1mm, right=1mm, top=0mm, bottom=0mm,
    before skip=0pt, after skip=0pt,
  },
   bubblepurple/.style={
    colback=lightpurple!20,
    colframe=lightpurple,
    boxrule=0.5mm,
    arc=2mm,
    left=1mm, right=1mm, top=0mm, bottom=0mm,
    before skip=0pt, after skip=0pt,
  }
}

\newsavebox{\roundimagebox}
\newlength{\roundimagewidth}
\newlength{\roundimageheight}

\newcommand{\roundimage}[2]{%
  \sbox{\roundimagebox}{%
    \includegraphics[width=#2\linewidth]{#1}%
  }%
  \setlength{\roundimagewidth}{\wd\roundimagebox}%
  \setlength{\roundimageheight}{\ht\roundimagebox}%
  \begin{tikzpicture}
    \clip[rounded corners=0.03\linewidth]
         (0,0) rectangle (\roundimagewidth, \roundimageheight);
    \node[anchor=south west, inner sep=0] at (0,0){%
      \usebox{\roundimagebox}%
    };
  \end{tikzpicture}%
}

\newcommand{\roundimageRight}[2]{%
  \begin{flushright}
    \sbox{\roundimagebox}{%
      \includegraphics[width=#2\linewidth]{#1}%
    }%
    \setlength{\roundimagewidth}{\wd\roundimagebox}%
    \setlength{\roundimageheight}{\ht\roundimagebox}%
    \begin{tikzpicture}
      \clip[rounded corners=0.03\linewidth]
           (0,0) rectangle (\roundimagewidth, \roundimageheight);
      \node[anchor=south west, inner sep=0] at (0,0){%
        \usebox{\roundimagebox}%
      };
    \end{tikzpicture}%
  \end{flushright}
}
\newcommand{\human}[1]{%
  \begin{flushleft}
    \begin{tcolorbox}[humanbubble, width=0.95\linewidth]
      \begin{minipage}[t]{0.1\linewidth}
        \vspace{-2pt}
        \includegraphics[scale=0.015]{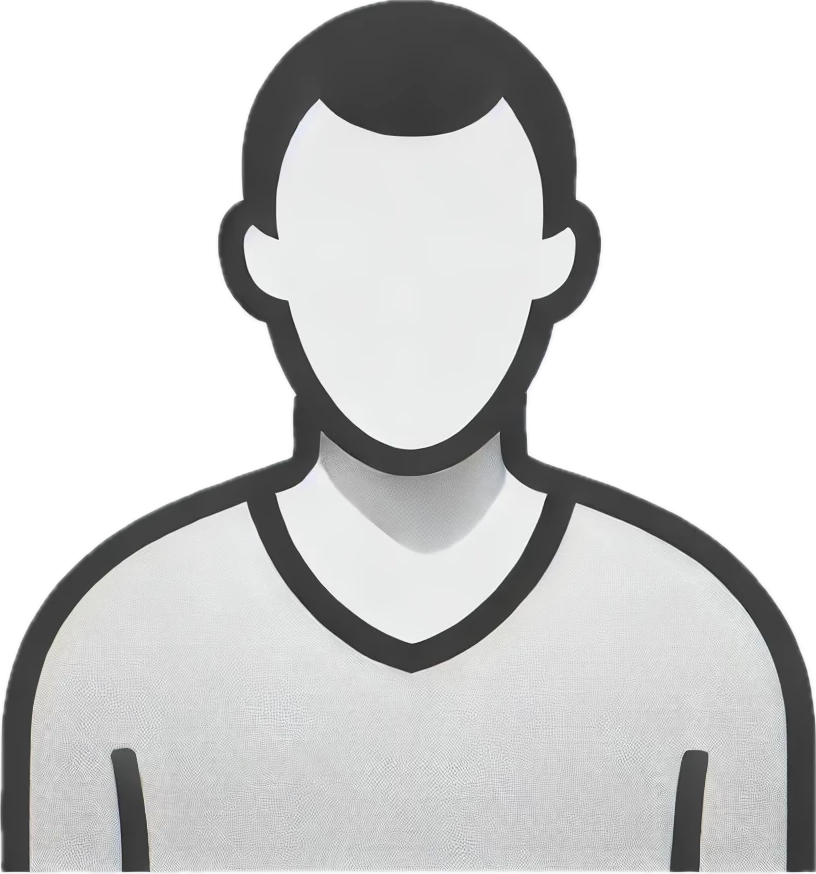}
      \end{minipage}%
      \begin{minipage}[t]{0.85\linewidth}
        \vspace{0pt}
        #1
      \end{minipage}
    \end{tcolorbox}
  \end{flushleft}
  \vspace{-6mm} 
}

\newcommand{\assistant}[1]{%
  \begin{flushright}
    \begin{tcolorbox}[assistantbubble, width=0.95\linewidth]
      \begin{minipage}[t]{0.9\linewidth}
        \vspace{0pt}
        #1
      \end{minipage}%
      \hspace{2mm}
      \begin{minipage}[t]{0.05\linewidth}
        \vspace{-2pt}
        \includegraphics[scale=0.015]{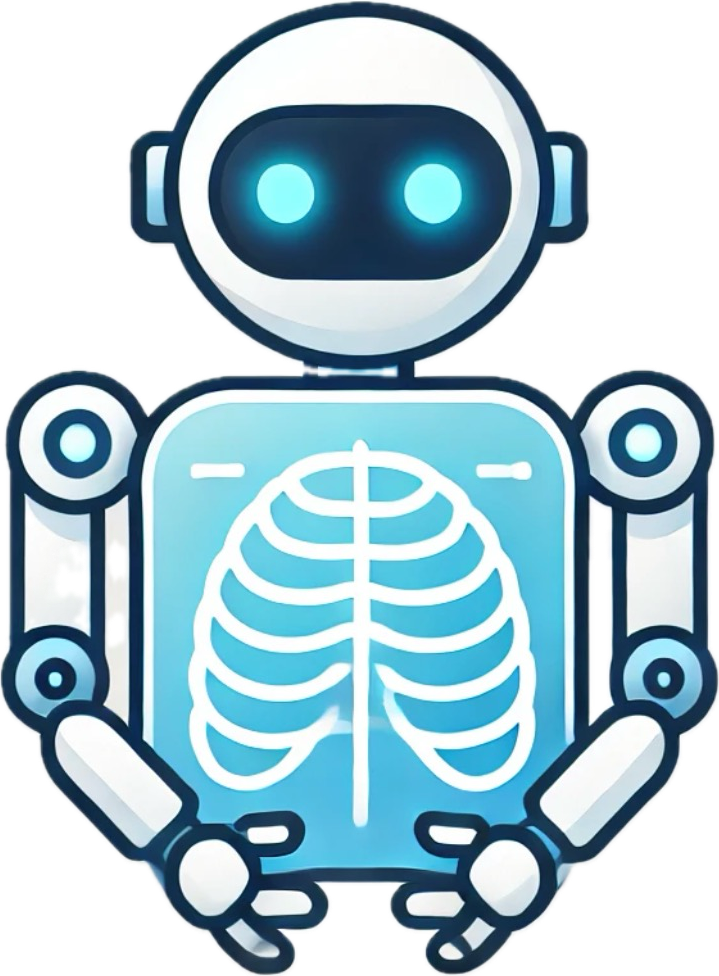}
      \end{minipage}
    \end{tcolorbox}
  \end{flushright}
  \vspace{-6mm}
}

\newcommand{\reference}[2][bubblered]{%
  \begin{flushright}
    \begin{tcolorbox}[#1, width=0.95\linewidth]
      \begin{minipage}[t]{0.9\linewidth}
        \vspace{0pt}
        #2
      \end{minipage}%
      \hspace{2mm}
      \begin{minipage}[t]{0.05\linewidth}
        \vspace{-2pt}
      \end{minipage}
    \end{tcolorbox}
  \end{flushright}
  \vspace{-6mm}
}

\newcommand{\radialog}[2][bubblebrown]{%
  \begin{flushright}
    \begin{tcolorbox}[#1, width=0.95\linewidth]
      \begin{minipage}[t]{0.9\linewidth}
        \vspace{0pt}
        #2
      \end{minipage}%
      \hspace{2mm}
      \begin{minipage}[t]{0.05\linewidth}
        \vspace{-2pt}
        RaDialog%
      \end{minipage}
    \end{tcolorbox}
  \end{flushright}
  \vspace{-6mm}
}

\newcommand{\llavamed}[2][bubblepurple]{%
  \begin{flushright}
    \begin{tcolorbox}[#1, width=0.95\linewidth]
      \begin{minipage}[t]{0.9\linewidth}
        \vspace{0pt}
        #2
      \end{minipage}%
      \hspace{2mm}
      \begin{minipage}[t]{0.05\linewidth}
        \vspace{-2pt}
        LLaVA-Med%
      \end{minipage}
    \end{tcolorbox}
  \end{flushright}
  \vspace{-6mm}
}

\newcommand{\regionimage}[2]{%
  \begin{minipage}[t]{0.23\linewidth} 
    \vspace{0pt}
    \centering
    \begin{tikzpicture}
      \clip[rounded corners=5pt] (0,0) rectangle (1.5cm,1.5cm);
      \includegraphics[width=1.5cm,height=1.5cm]{#1}
    \end{tikzpicture}\\[-1mm]
    {\scriptsize #2}
  \end{minipage}%
}

\title{RadVLM: A Multitask Conversational Vision-Language Model for Radiology}

\date{}
\author[1\footnote{Correspondence: nicolas.deperrois@uzh.ch}]{Nicolas Deperrois}
\author[2]{Hidetoshi Matsuo}
\author[3]{Samuel Ruipérez-Campillo}
\author[3]{Moritz Vandenhirtz}
\author[3]{Sonia Laguna}
\author[3]{Alain Ryser}
\author[4]{Koji Fujimoto}
\author[2]{Mizuho Nishio}
\author[3]{Thomas M. Sutter}
\author[3]{Julia E. Vogt}
\author[1,5]{Jonas Kluckert}
\author[5]{Thomas Frauenfelder}
\author[5]{Christian Bl\"uthgen}
\author[1\footnote{Joint senior authorship.}]{Farhad Nooralahzadeh}
\author[1$^\dagger$]{Michael Krauthammer}

\affil[1]{Department of Quantitative Biomedicine, University of Zurich, Zurich, Switzerland}
\affil[2]{Department of Radiology, Kobe University, Kobe, Japan}
\affil[3]{Department of Computer Science, ETH Zurich, Zurich, Switzerland}
\affil[4]{Department of Advanced Imaging in Medical Magnetic Resonance, Kyoto University, Kyoto, Japan}
\affil[5]{Diagnostic and Interventional Radiology, University Hospital Zurich, Zurich, Switzerland}

\begin{document}

\maketitle

\begin{abstract}
The widespread use of chest X-rays (CXRs), coupled with a shortage of radiologists, has driven growing interest in automated CXR analysis and AI-assisted reporting. While existing vision-language models (VLMs) show promise in specific tasks such as report generation or abnormality detection, they often lack support for interactive diagnostic capabilities. 
In this work we present RadVLM, a compact, multitask conversational foundation model designed for CXR interpretation. To this end, we curate a large-scale instruction dataset comprising over 1 million image-instruction pairs containing both single-turn tasks -- such as report generation, abnormality classification, and visual grounding -- and multi-turn, multi-task conversational interactions. After fine-tuning RadVLM on this instruction dataset, we evaluate it across different tasks along with re-implemented baseline VLMs. 
Our results show that RadVLM achieves state-of-the-art performance in conversational capabilities and visual grounding while remaining competitive in other radiology tasks. Ablation studies further highlight the benefit of joint training across multiple tasks, particularly for scenarios with limited annotated data.
Together, these findings highlight the potential of RadVLM as a clinically relevant AI assistant, providing structured CXR interpretation and conversational capabilities to support more effective and accessible diagnostic workflows\footnote{RadVLM's instruction dataset is available at \href{https://physionet.org/content/radvlm-instruction-dataset}{https://physionet.org/content/radvlm-instruction-dataset} and the model weights are available at \href{https://physionet.org/content/radvlm-model}{https://physionet.org/content/radvlm-model}}.
\end{abstract}

\section{Introduction}
X-rays have played a fundamental role in medicine since their discovery in 1895~\citep{rontgen1895ueber}, and continue to be the most frequently used medical imaging modality worldwide due to their convenience and cost-effectiveness \citep{akhter2023ai}. Chest X-ray (CXR) remains the most commonly performed radiological exam globally, particularly important for diagnosing and monitoring thoracic conditions such as pneumonia, heart failure, and lung cancer \citep{ccalli2021deep}. Problematically, the growing volume of CXRs and other imaging studies in recent years have led to a reduction in the time available for radiologists to thoroughly evaluate each case ~\citep{peng2022radiologist}. As a result, in many countries, the responsibility of interpreting CXRs is often transferred to non-radiology physicians, who typically possess less specialized training and experience. This shift increases the risk of diagnostic errors or misinterpretations \citep{Malak2021, peng2022radiologist}.

The shortage of trained personnel for CXR interpretation has led to the exploration of automated agents to assist physicians in diagnostic tasks. In recent years, various deep learning models have shown promise in clinical applications, such as the detection of conditions like COVID-19 pneumonia \citep{Nishio2020} or pulmonary nodules \citep{Homayounieh2021}. Another extensively studied task is the automated generation of free text reports from CXR images using transformer-based architectures \citep{nooralahzadeh2021progressive, Yang2023, hyland2023maira, chaves2024llavarad}. These models can provide preliminary drafts summarizing key observations from the CXR, offering a potential enhancement to the diagnostic workflow. There is a need to expand the scope of these tools beyond report generation, towards the capability to answer questions about the CXR technique, findings in a region of interest, location of specific abnormalities, and definitions of medical terms. In addition, physicians should be allowed to formulate their queries flexibly and in any order, potentially within a multi-turn, conversational interaction with the assistant \citep{Tu2024-ad}.

Recently, significant advancements in the field of multimodal artificial intelligence (AI) have enabled the development of models such as GPT-4 Vision  \citep[GPT4-V,][]{chatgpt-ut} and Claude \citep{claude-hq}, which have the ability to describe and converse about images with increasing reliability. The training principle behind these models relies on an initial stage to align pretrained vision and language modules, followed by a visual instruction tuning phase where the model learns to respond to image-related queries and commands \citep{li2024llava}. Such advancements have inspired the adaptation of multimodal AI assistants for medical applications \citep{singhal2023large, li2023llava-med, saab2024capabilities}. 

Despite these advancements, there remains a need for specialized multimodal conversational assistants tailored specifically for CXR interpretation. In this direction, models such as CheXagent \citep{chen2024chexagent}, RaDialog \citep{pellegrini2025radialog}, or MAIRA-2 \citep{Bannur2024-ek} were developed, extending beyond report generation to tasks such as observation grounding and visual question answering, covering a larger part of the clinical workflow. However, their capacity to handle diverse and complex user queries, or to respond accurately to multiple prompts within an arbitrary conversational framework, remains limited. Adding these capabilities is critical for comprehensively supporting  clinicians' daily work.

In this study, we build upon state-of-the-art visual instruction-tuning techniques inspired by general-domain applications \citep{liu2023visual, wang2024qwen2vl} to construct a compact, multitask conversational foundation model specialized in CXR interpretation. To achieve this aim, we create comprehensive CXR datasets, each featuring diverse modalities including free-text reports, abnormality labels, and visual coordinates, and organize them into a unified instruction dataset. This dataset is comprised of single-turn image-instruction pairs for different tasks and image-conversation pairs designed for more flexible and multi-turn interactions. We then fine-tune a vision-language architecture \citep{li2024llava} on this instruction dataset, naming the resulting model RadVLM, and develop an evaluation pipeline to assess its performance across multiple tasks, systematically comparing it to state-of-the-art generalist and CXR-specific foundation models. Our results show that, despite its relatively compact size, RadVLM achieves competitive performance on individual tasks relevant to clinical practice, providing conversational capabilities within a simple and flexible interface, providing a reliable and user-friendly tool for physicians. Additionally, we compare RadVLM with several existing vision-language models and show that RadVLM matches or outperforms these across both individual and conversational tasks. 

In summary, the contributions of this work are as follows: 
\begin{itemize}
    \item We develop a unique instruction dataset that extends beyond report generation to encompass diverse CXR-based tasks, including multi-turn conversational interactions tailored for clinical workflows.
    \item We design and train RadVLM, a multitask conversational foundation model designed to assist physicians in CXR analysis that solely relies on visual information -- avoiding the need for providing additional metadata. 
    \item We employ an evaluation pipeline, re-implementing existing models for comparison and ensuring reproducibility of results.
    \item We evaluate RadVLM systematically across multiple tasks, demonstrating competitive performance against state-of-the-art vision-language models, both generalist, and medical-specific.  In particular, we evaluate conversational abilities in clinical contexts and demonstrate that RadVLM significantly  outperforms existing general and clinical VLMs in this aspect.
\end{itemize}

\section{Related work}

\subsection{Instruction tuning and vision-language models}

The advent of autoregressive large language models (LLMs) based on the transformer architecture ~\citep{vaswani2017attention} and pre-trained on vast text corpora~\citep{radford2019language, brown2020language} has provided the possibility to perform a wide range of language-based downstream tasks. However, the widespread success and accessibility of LLMs, such as ChatGPT, are largely attributed to the instruction-tuning process \citep{wei2021finetuned, ouyang2022training}. This process commonly involves fine-tuning a pre-trained model on a labeled dataset of diverse instruction-following tasks, ensuring the model can generalize to diverse user instructions in a zero-shot setting. 

Instruction-following datasets generally consist of instruction-output pairs and/or multi-turn dialogues \citep{zheng2023lmsys} mimicking real-life interaction between users and AI assistants. While early instruction datasets were manually crafted \citep{wei2021finetuned}, a more scalable approach leverages larger LLMs to generate synthetic instruction data \citep{wang2022self, peng2023instruction, liu2023visual}, reducing annotation costs. 

Beyond the text-only tasks, state-of-the-art proprietary LLMs, such as GPT-4 \citep{achiam2023gpt}, DeepSeek \citep{liu2024deepseek, guo2025deepseek}, and Gemini \citep{team2023gemini} exhibit advanced vision capabilities, enabling them to process and respond to multimodal instructions. In parallel, open research efforts have led to the development of vision-language models such as LLaVA \citep{liu2023visual} and BLIP-2 \citep{li2023blip}, which introduced effective training strategies for visual instruction tuning. These approaches have inspired the development of vision-language models (VLMs) such as LLaVA-OneVision \citep{li2024llava}, Idefics3 \citep{laurencon2024building}, Qwen2-VL \citep{wang2024qwen2vl}, and Llama-3.2 Vision \citep{dubey2024llama}. Similar to text-based LLMs, the instruction-following datasets contain user–assistant Q\&A and dialogues, but each example is paired with an image, and the instructions and responses explicitly reflect the image's content \citep{feng2022mmdialog}.

\subsection{Vision-language models in radiology}

 The success of VLMs in the general domain has spurred the development of medical-based VLMs, particularly in domains where image-based interpretation is critical. Proprietary models such as Med-PaLM\citep{singhal2023large} and Med-Gemini \citep{saab2024capabilities} have shown remarkable performance across a range of multimodal medical tasks, including medical visual question answering (VQA), report generation, summarization. In parallel, open source models such as LLaVA-Med \citep{li2023llava-med} have been developed following similar training strategies as LLaVA \citep{liu2023visual}, leveraging biomedical datasets from PubMed \citep{pubmed} to design instruction prompts and muti-turn conversations. 

Among medical applications, CXR interpretation remains a key area of interest. Early AI-driven models primarily focus on report generation \citep{nooralahzadeh2021progressive, alfarghaly2021automated, tanida2023interactive, chaves2024llavarad}, supported by the development of clinically relevant evaluation metrics \citep{jain2021radgraph, yu2023evaluating}. More recently, research has expanded toward multimodal, multitask CXR assistants capable of integrating multiple functionalities beyond report generation, such as classification, grounding or image generation. Notable examples include CheXagent \citep{chen2024chexagent} or RoentGen \citep{bluethgen2024vision}, though these models lack conversational capabilities. 

Other approaches, such as Wolf \citep{kang2024wolf}, RaDialog  \citep{pellegrini2025radialog}, and M4CXR \citep{park2024m4cxr}, incorporate conversational features but are constrained by predefined response templates, limiting their adaptability in real-world interactions. In this work, we introduce a model that integrates multiple CXR interpretation tasks while enabling flexible,  multi-turn dialogue, bridging the gap between task-specific AI models and interactive clinical assistants.

\section{Methods}

\subsection{Instruction dataset}

A key step in the development of RadVLM is the construction of an instruction dataset. For this purpose, we first aggregate and process multiple publicly available datasets containing CXR images paired with various attributes, including free-text reports, categorical labels, and bounding boxes. From these sources, we generate a dataset of over 1 million instruction instances, each consisting of a frontal CXR image and a corresponding user-assistant interaction derived from the available attributes. These interactions can be in the form of a single Q\&A designed for predefined tasks (single instructions) or of a multi-turn exchange (conversations). The composition of this instruction dataset is detailed below and summarized in \cref{table:instruction-dataset-content}. 

\subsubsection{Free-text report generation}

In alignment with existing CXR models, we aim to generate clinically coherent radiology reports from CXR images. To achieve this, we collect public datasets containing CXRs paired with anonymized free-text reports. We remove the clinical and indication sections from each report (indication is not used in training) and keep the findings section as the target; if the findings section is empty, we instead use the impression section, which often contains the relevant findings instead. We also filter out lateral CXR images from training, as frontal views contain the most essential diagnostic information.  

Radiology reports often compare to and refer to prior X-ray examinations when discussing current radiological findings. These earlier images should be provided as part of the prompt when conducting report generation based on raw radiology reports \citep{kim2023boosting, Bannur2024-ek}. As we focus on the analysis of a single CXR image, we therefore used GPT-4o through a secure Azure OpenAI service -- configured per the guidance outlined in the relevant PhysioNet statement\footnote{\url{https://physionet.org/news/post/gpt-responsible-use}} -- to remove mentions of prior studies, consistent with recent work on report generation \citep{chen2024chexagent, chaves2024llavarad}.

We leverage two public datasets for the report generation task:

\noindent \textbf{MIMIC-CXR} \citep{johnson2019mimic}, which contains 377,110 CXR image-report pairs from 227,943 studies. After filtering, we retain 230,980 image-text pairs in the training set and 3,314 in the test set.

\noindent \textbf{CheXpert-Plus} \citep{chambon2024chexpertplusaugmentinglarge}, which features 223,228 image-report pairs from 187,711 studies. Applying the same filtering process as for MIMIC-CXR, we obtain 186,463 image-text training pairs.

\begin{figure}[t]
\centering 
{%
\footnotesize 
\begin{minipage}[t]{0.48\linewidth} 
  \vspace{0pt} 
  \textbf{a. Report generation}
  \human{Draft a concise report for this image.
  \vspace{1mm} 
  
  \roundimage{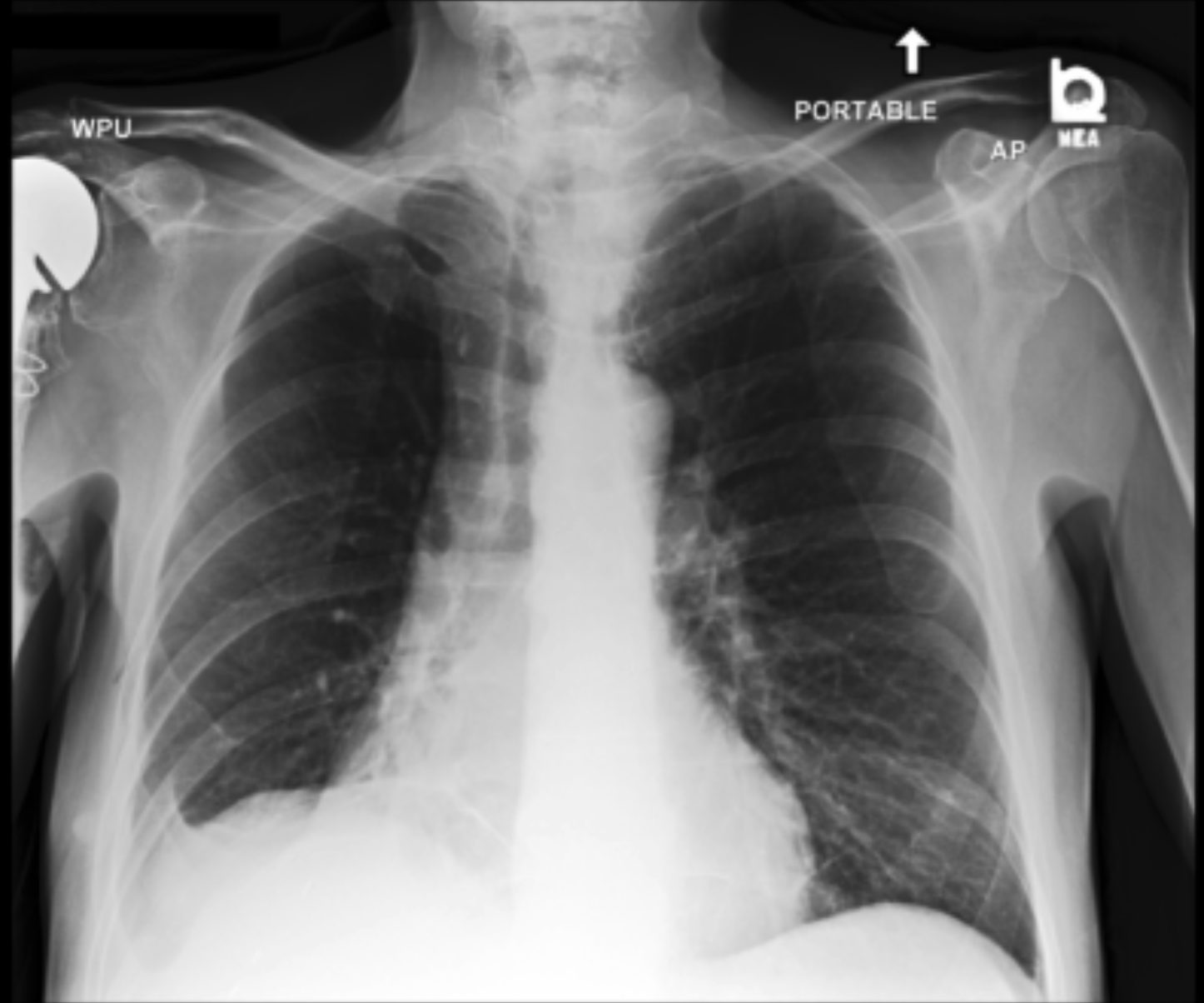}{0.2}
  }
  \reference{Small right pleural effusion is present. Minimal right lower lobe atelectasis is likely. The right lung is otherwise clear. No appreciable pneumothorax is seen. Biapical margin thickening, left worse than right, is seen. The left lung is well-expanded and clear. The heart size is normal. Hemi arthroplasty is seen on the right shoulder.}
  \vspace{5mm}
  \centering
\end{minipage}%
\hfill
\begin{minipage}[t]{0.48\linewidth} 
  \vspace{0pt} 
  \textbf{b. Abnormality classification}
  \human{Are there identifiable abnormalities on this Chest X-ray?
  
  \vspace{1mm} 
  \roundimage{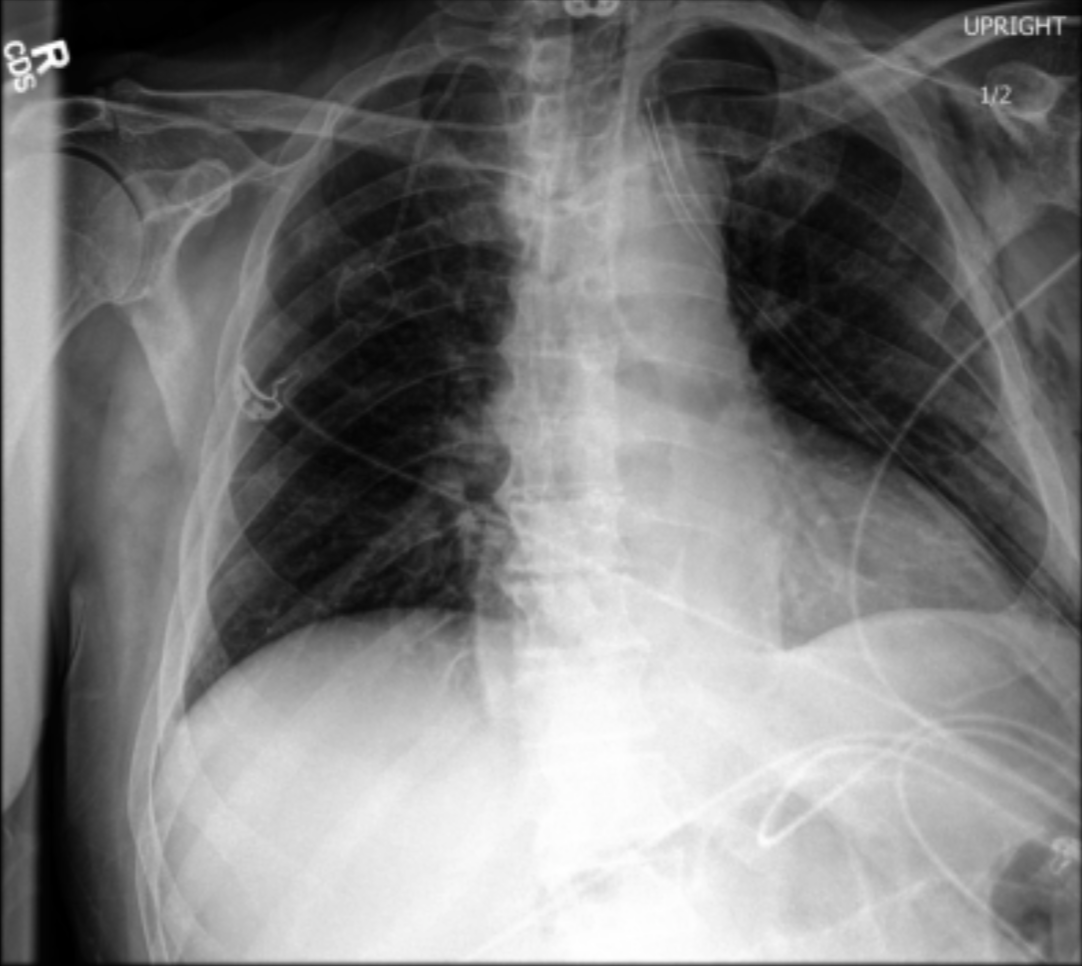}{0.2}
  }
  \reference{Yes, there is one abnormality identified: pneumothorax.}
  \vspace{5mm}
  \centering
\end{minipage}%
\vspace{2mm} 

\begin{minipage}[t]{0.48\linewidth} 
  \vspace{0pt} 
  \textbf{c. Visual grounding: Anatomy/Abnormality}
   \human{Indicate the position of the \textit{left apical zone}.
  \vspace{1mm} 
  \roundimage{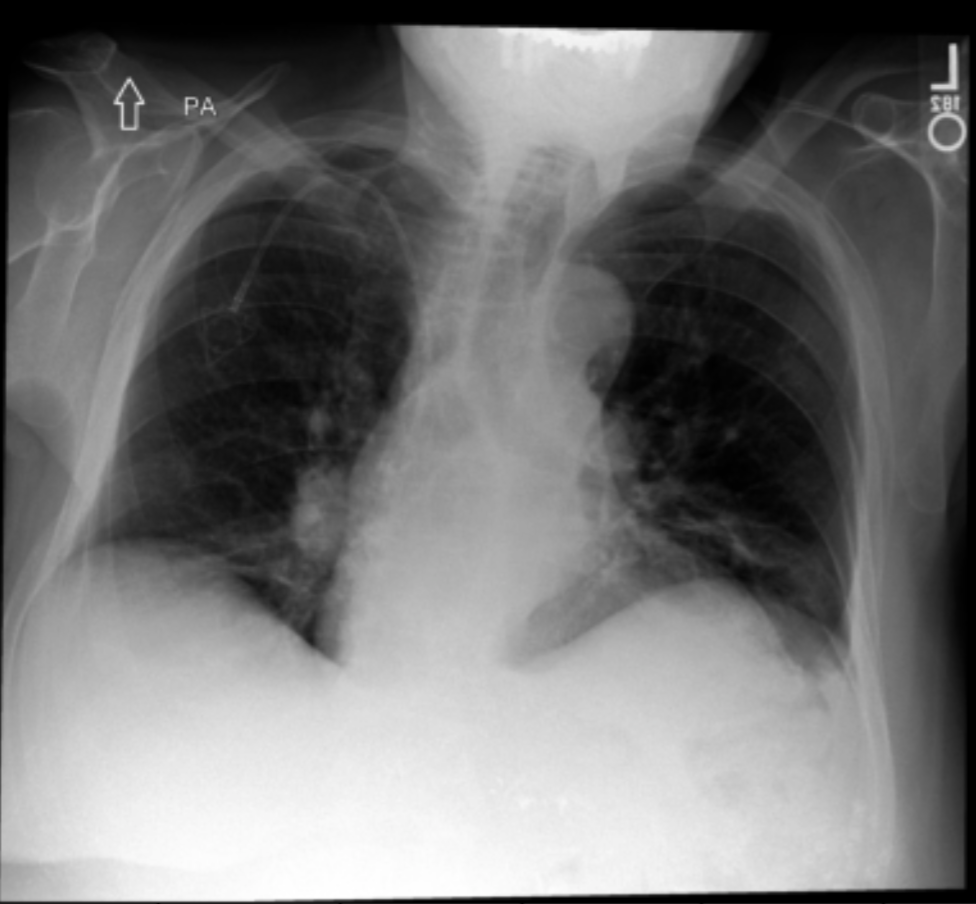}{0.2}
  }
  \reference{The left apical zone is situated at [0.57, 0.14, 0.85, 0.3] in the image.
  \roundimageRight{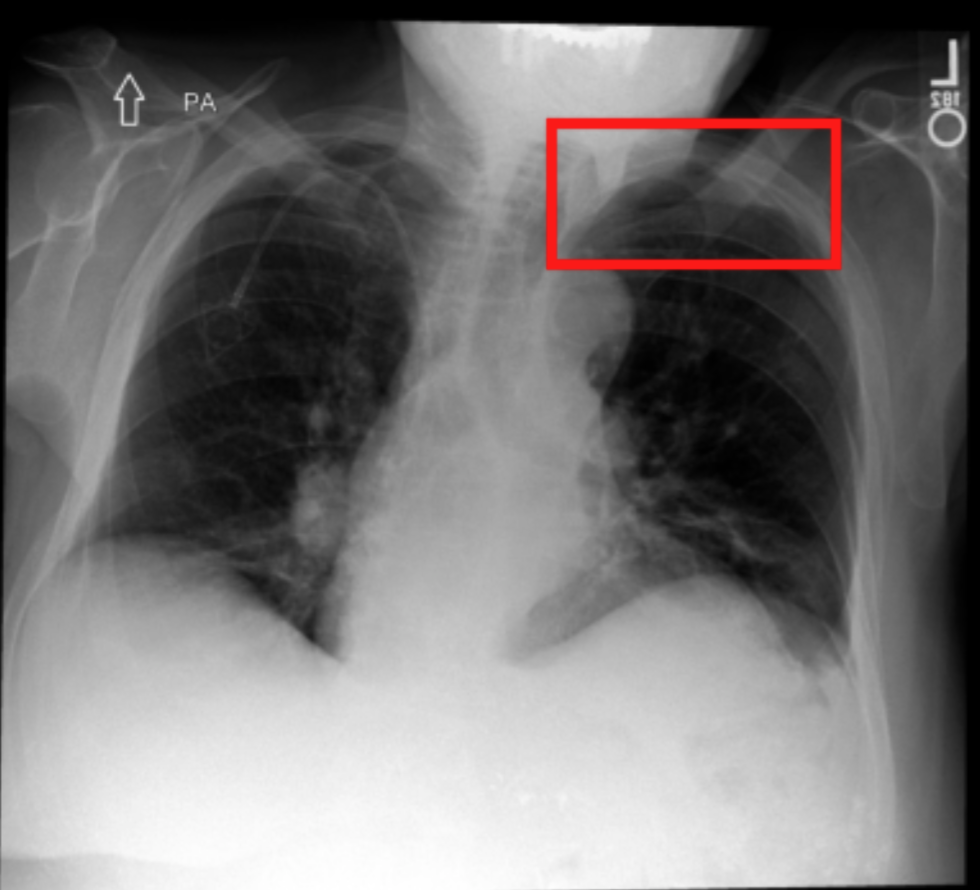}{0.2}
  }
  \vspace{5mm}
  \centering
\end{minipage}%
\hfill
\begin{minipage}[t]{0.48\linewidth} 
  \vspace{0pt} 
  \textbf{d. Visual grounding: Phrase}
   \human{Mark the area where you observe: \textit{marked elongation and increased caliber of the thoracic aorta}.
   
  \vspace{1mm} 
  \roundimage{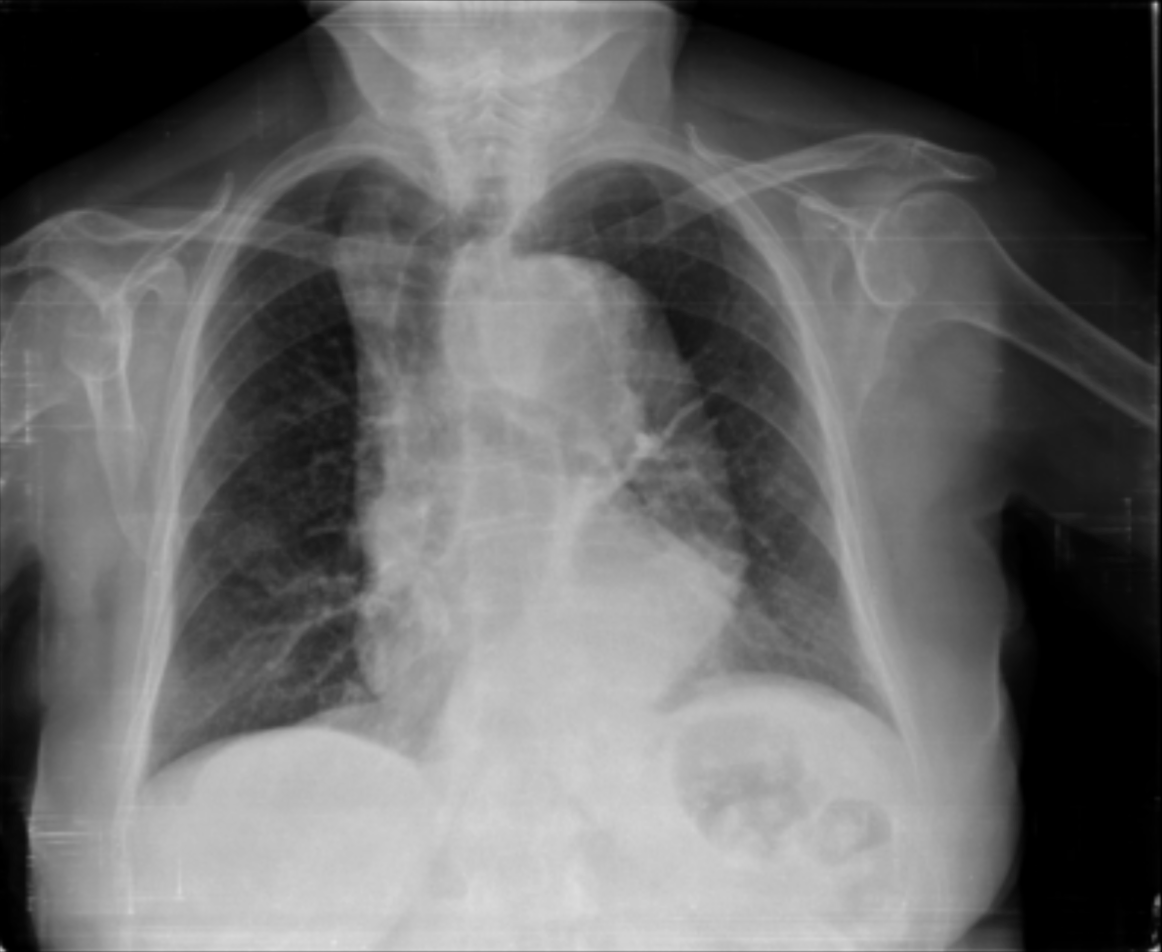}{0.22}
  }
  \reference{The area specified is at coordinates [0.39, 0.3, 0.66, 0.79].
  \roundimageRight{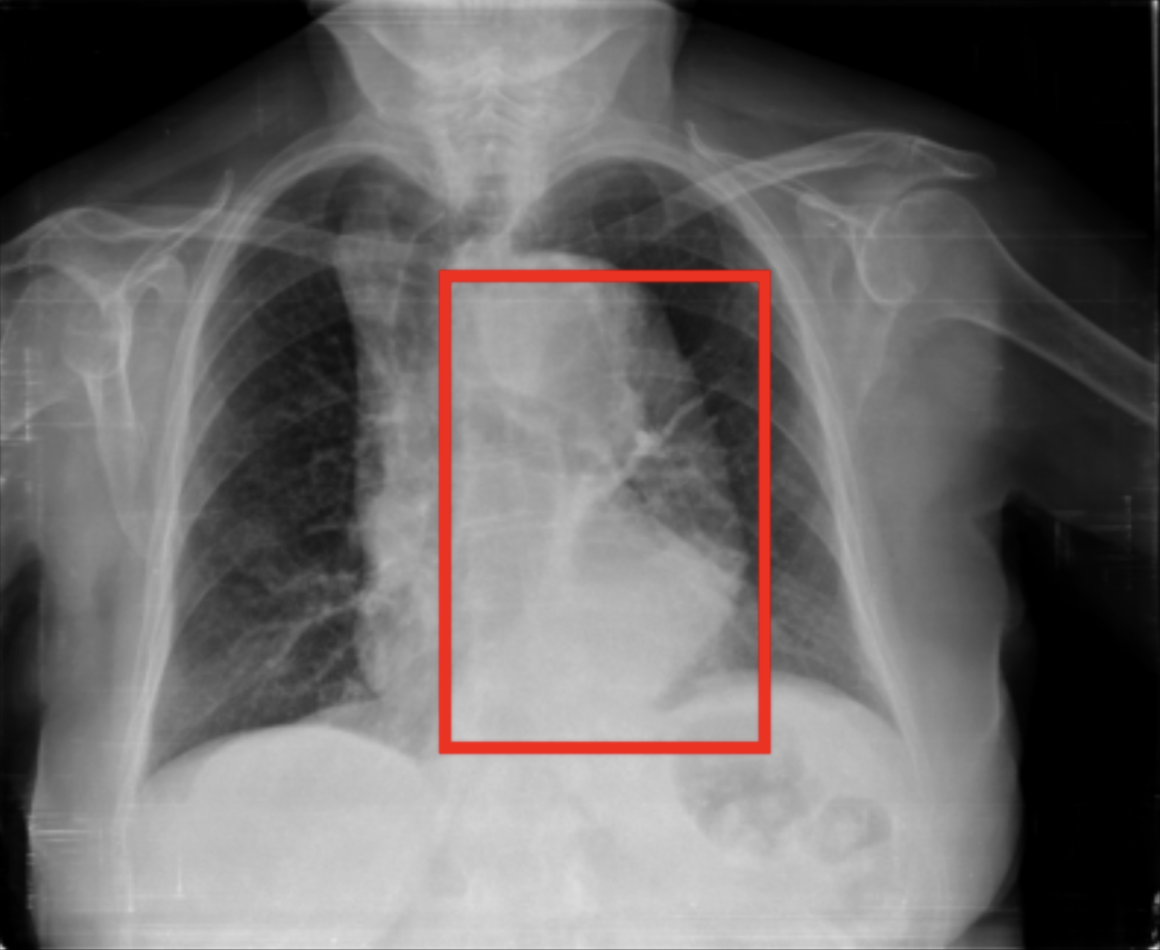}{0.22}
  }
   \vspace{5mm}
  \centering
\end{minipage}%
}
\caption{ \textbf{Examples of single instructions for different tasks.} We design three main types of instructions based on dataset attributes. For datasets containing image-reports pairs (e.g., MIMIC-CXR), we design Q\&A towards report generation (a). The instructions for datasets containing abnormality labels (e.g., CheXpert) are designed to perform multi-class classification (b). When bounding boxes are available, we design visual grounding instructions, where the assistant provides the bounding box coordinates to display them on the input image (c,d). }
\label{fig:single-instructions}
\end{figure}

For this task, the instructions are designed such that the user asks the assistant to generate a report for a given CXR, and the assistant responds with the filtered findings corresponding to that CXR (\cref{fig:single-instructions}a).

\subsubsection{Abnormality classification}

Another essential task an AI-assistant should be capable of is to identify the presence of abnormalities on a CXR. While simpler than generating detailed, unstructured observations, this functionality serves as a quick and helpful overview for physicians, highlighting key observations before they dive into a more detailed analysis.

For this task, we collect CXR datasets paired with abnormality labels. These labels were extracted from the original textual reports via the CheXbert automatic labeling tool \citep{smit2020chexbert},  which identifies whether each of 14 possible abnormalities is present, absent, or uncertain. In our setup, we only consider frontal images and--in line with previous work \citep{yang2024advancing, chaves2024llavarad}--consider ``uncertain'' abnormalities as ``absent''.

For this task, we use 191,027 image-labels pairs from CheXpert \citep{irvin2019chexpert} and 237,912 pairs from MIMIC-CXR (\cref{table:instruction-dataset-content}). The instructions are designed such that the user asks for the abnormalities present on the CXR and the assistant answers by providing the list of abnormalities (\cref{fig:single-instructions}b). 

\subsubsection{Visual grounding}

Detecting the location of specific anatomical regions or pathologies on a CXR is an important task for AI assistants. In addition to providing a textual description of the image, they should be able to spot where specific observations are located. This is usually done by predicting bounding box coordinates of top-left and bottom-right corners $[x_1, y_1, x_2, y_2]$. While classical object detectors \citep{ren2016faster, redmon2016you} tackle this task by leveraging specialized architectures and learning rules, we embark on it with the other text-based tasks via next-token prediction (\cref{eqn:next-word-prediction}), formatting coordinates into text, enclosed between square brackets. As input images are pre-processed to a unique size by the vision encoder, we normalize these coordinates to the original dimension to obtain floating values between 0 and 1, similarly as in \citet{you2023ferret, park2024m4cxr, zhang2024ferret}.

The following datasets, with fine-grained  X-ray information, were collected to design visual grounding instructions: 

\noindent
\textbf{Chest Imagenome} \citep{wu2021chest}, derived from MIMIC-CXR, provides additional annotations to frontal X-ray images, in particular, bounding box coordinates for 29 anatomical regions. In our training data, we randomly select one region per image for each datapoint and create an instruction for anatomical region grounding following \cref{fig:single-instructions}. 

\noindent
\textbf{VinDr-CXR} \citep{nguyen2022vindr} contains 18,000 frontal images, each manually annotated by three different radiologists. To merge their annotations, we pre-processed them by fusing bounding boxes of the same pathology using weighted box fusion \citep{solovyev2021weighted}, similarly as in \citet{muller2024chex}. From this dataset, we design two types of tasks: i) abnormality grounding, asking for the location of a specific abnormality (also following \cref{fig:single-instructions}c) and ii) abnormality detection, asking the location of all abnormalities, if any (\cref{table:instruction-dataset-content}). 

\noindent
\textbf{MS-CXR} \citep{boecking2022making} provides image-sentence pairs of bounding boxes and corresponding phrases, complementing MIMIC-CXR. 

\noindent
\textbf{PadChest-GR} \citep{castro2024padchest} also contains grounded sentences derived from the PadChest dataset \citep{bustos2020padchest}.  

From the last two datasets, we construct the `phrase grounding' task, where a user asks about the location of a specific sentence from a radiology report, and the assistant provides its associated bounding box coordinates (\cref{fig:single-instructions}d).

\begin{table}[t]
\centering
\footnotesize
\renewcommand{\arraystretch}{1.5} 
\begin{tabular}{p{4cm}p{2.5cm}C{4.5cm}C{2.8cm}}
\toprule
\multicolumn{1}{c}{\textbf{Task}} & \multicolumn{1}{c}{\textbf{Dataset source}} & \textbf{Image-instruction pairs (\#)} & \textbf{Evaluation (\#)} \\ \midrule
\multirow{2}{*}{Report Generation}
& MIMIC-CXR & 230,980 $\times$ 1 &  3,314 \\ 
& CheXpert-Plus & 186,463 $\times$ 1 &  - \\ \midrule
\multirow{2}{*}{Abnormality classif.}
 & MIMIC-CXR     & 237,912 $\times$ 1 &  518 \\ 
 & CheXpert     & 191,027 $\times$ 1 &  - \\ \midrule
\multirow{1}{*}{Anatomical grounding}
 & Chest Imagenome & 80,000 $\times$ 1 &  2,000 \\ \midrule
\multirow{1}{*}{Abnormality grounding}
& VinDr-CXR  & 16,089 $\times$ 3 &  2,108 \\ \midrule
\multirow{1}{*}{Abnormality detection}
& VinDr-CXR  & 15,000 $\times$ 2 &  - \\ \midrule
\multirow{2}{*}{Phrase grounding}
& MS-CXR  & 971 $\times$ 3 &  189 \\
 & PadChest-GR & 4478 $\times$ 2 & -  \\ \midrule
\multirow{1}{*}{Conversation}
 & MIMIC-CXR & 86,155 $\times$ 1 &  500 \\ \midrule
\multirow{2}{*}{Conversation (grounded)}
 & MS-CXR & 862 $\times$ 4 &  155 \\
& PadChest-GR & 2,225 $\times$ 4 & - \\ 
\bottomrule
\end{tabular}
\caption{\textbf{Overview of the instruction dataset.} The instruction dataset comprises 1,115,021 image-instruction pairs spanning multiple vision-language tasks, including report generation, abnormality classification, anatomical and abnormality grounding, phrase grounding, and conversational interactions. Dataset sources and the corresponding number of image-instruction pairs are listed, with smaller datasets balanced by varying the frequency of instruction occurrences.}
\label{table:instruction-dataset-content}
\end{table}

\subsubsection{Conversations}
\label{section:conversation}

Fine-tuning a VLM on single instructions, as presented above, is useful to acquire maximal precision in specific tasks but would not be sufficient to build a robust, flexible, and conversational radiology assistant. First, in a real-life setting, we cannot assume that physicians prompt the model with a limited set of instructions. Various types of questions could be posed, such as asking about the characteristics of a specific organ (lungs, heart), the orientation of the X-ray, or the definition of certain medical terms. More importantly, interactions can be decomposed over several Q\&A rounds, with sometimes a question referring to previous answers (e.g., asking about the location of a specific observation from the previous answer). The model should thus be tuned to sequentially connect visual concepts (textual observations, presence or absence of abnormalities, fine-grained information) throughout a single conversation thread.

\begin{figure}[t!]
\footnotesize 
\begin{minipage}[t]{0.48\linewidth} 
\vspace{0pt} 
  \textbf{a. Sample conversation}
  \human{Can you describe what you see in the chest X-ray?
  \vspace{1mm} 
  
  \roundimage{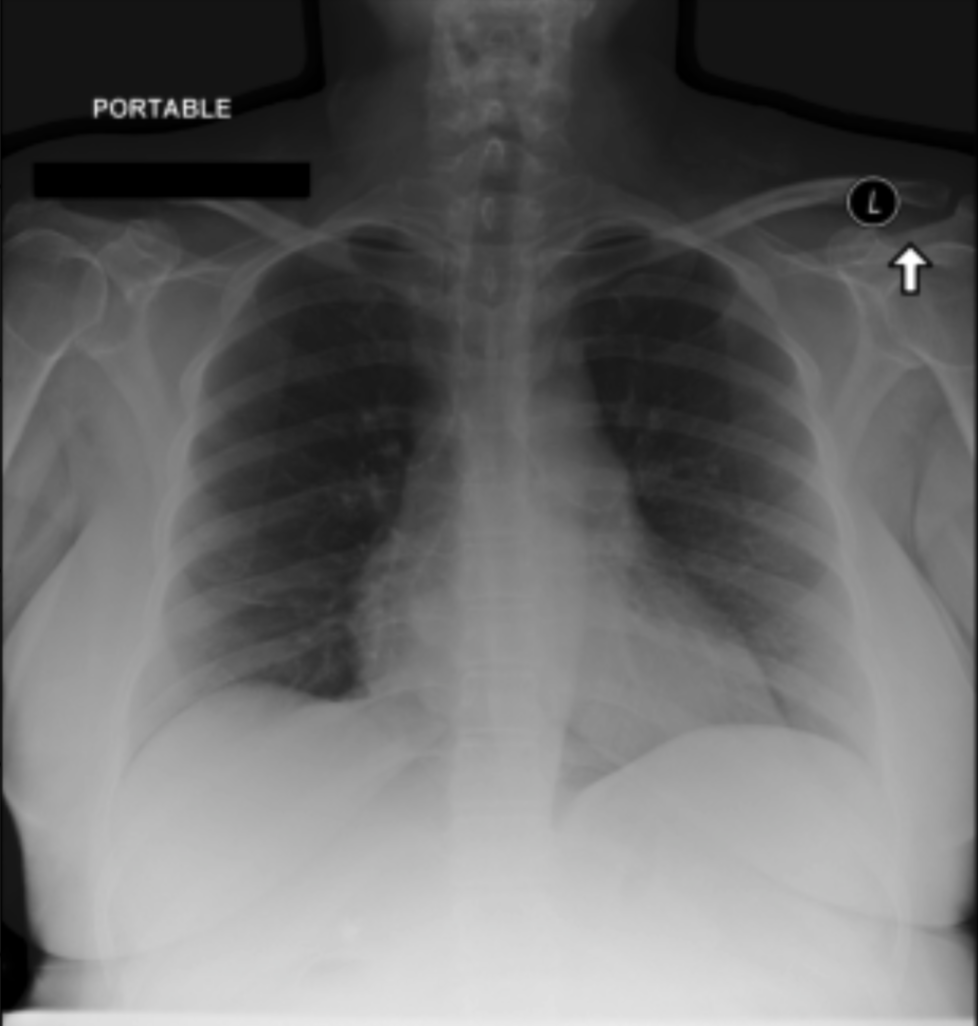}{0.2}
  }
  \reference{Sure, I see a chest X-ray taken from the front view, known as AP (anterior-posterior). The heart appears to be borderline enlarged, but otherwise, the mediastinal and hilar regions look normal. There are no signs of fluid around the lungs or collapsed lung areas.}
  \human{Is there anything unusual in the lung area?}
  \reference{The lungs mostly look clear, but there is some linear scarring or atelectasis in the right lung base. Apart from that, the lung fields do not show any significant abnormalities.}
  \human{Are there any signs of past surgery visible in the X-ray?}
  \reference{Yes, there are clips visible in the right upper quadrant of the abdomen, which are often left after a cholecystectomy, a surgical procedure to remove the gallbladder.}
  \human{Does the X-ray show any signs of fluid around the lungs?}
  \reference{No, the X-ray does not show any signs of pleural effusion, which is fluid accumulation around the lungs.}
  \human{Is the heart looking normal in size and shape?}
  \reference{The heart size appears borderline enlarged, which means it is slightly bigger than expected, but not significantly abnormal in shape.}
\end{minipage}%
\hfill
\begin{minipage}[t]{0.48\linewidth} 
\vspace{0pt} 
  \textbf{b. Sample conversation with grounding}
\human{Can you tell me what you observe in this chest X-ray?
\vspace{1mm}

\roundimage{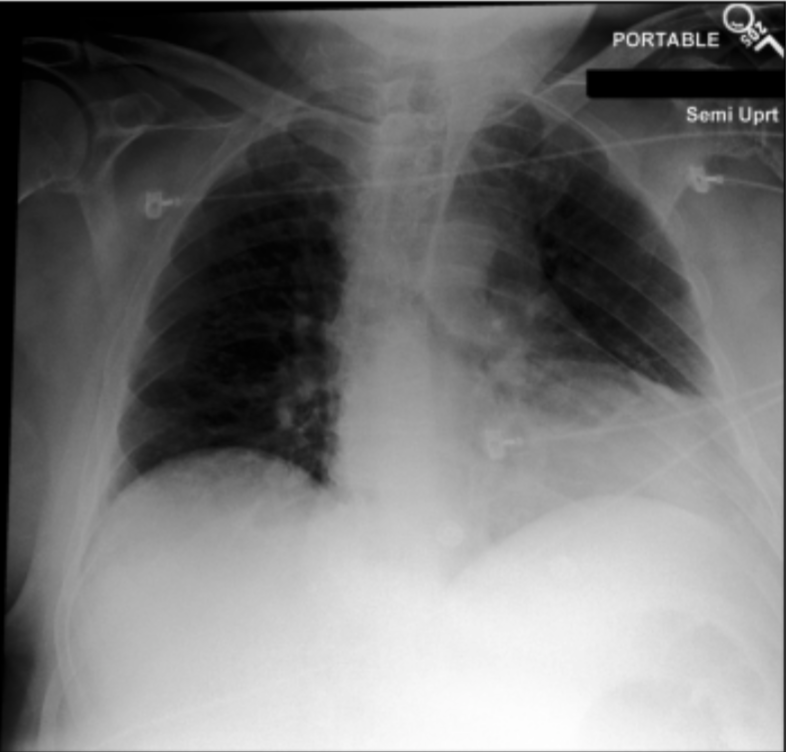}{0.2}
}
 \reference{The X-ray shows that the cardiac silhouette is enlarged. There is also suspicion of mild left basal atelectasis.}
 \human{Where is the cardiac silhouette enlargement located on the image?}
 \reference{The enlarged cardiac silhouette is located within the coordinates [0.51, 0.43, 0.98, 0.79].
  \roundimageRight{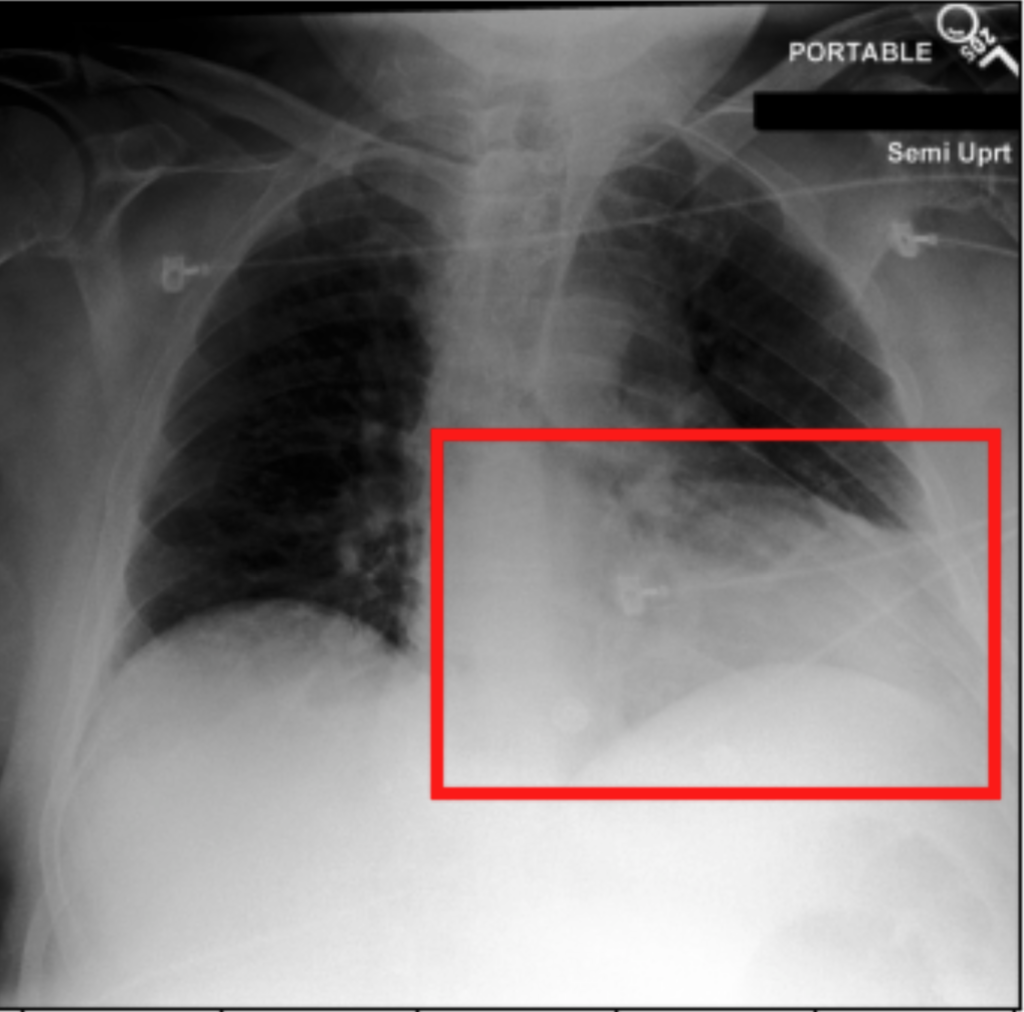}{0.2}
  }
  \human{Is there any fluid accumulation in the pleural space visible in the X-ray?}
  \reference{No, there is no visible pleural effusion on the X-ray.}
  \human{What can be said about the right lung in this image?}
  \reference{The right lung appears clear with no visible abnormalities.}
  \human{What does atelectasis mean?}
  \reference{Atelectasis refers to the partial collapse or incomplete inflation of the lung. In this X-ray, mild left basal atelectasis is suspected at the base of the left lung.}
\end{minipage}
\vspace{2mm}
\caption{ \textbf{Example of LLM-generated conversations within the instruction dataset}. LLM-generated user-assistant interactions designed for instruction tuning in RadVLM, covering both standard conversations and grounded responses.  \textbf{(a) Standard conversation}: The assistant responds to user queries based on textual attributes extracted from the CXR (e.g., report findings, categorical labels) without explicit spatial references. \textbf{(b) Conversation with grounding:} In addition to textual responses, the assistant provides spatial grounding by referencing anatomical structures with bounding box coordinates. These synthetic interactions are generated by conditioning a text-based LLM on CXR attributes (report, labels, bounding boxes) and prompting it to simulate multi-turn diagnostic dialogues.}
\label{fig:conversation}
\end{figure}

To develop this capability in RadVLM, we constructed an instruction-tuning dataset mimicking a real-life multi-turn interaction between user and assistant, named ``conversation dataset''. Here, questions can be asked in different order, and the assistant reacts to the content of previous answers. Inspired by the vision-language models LLaVA \citep{liu2023visual} and LLaVA-Med \citep{li2023llava-med}, we prompt a larger text-only LLM (GPT-4o) to generate multi-turn conversations. The prompt includes a system message instructing the LLM to generate a dialogue between a user and an assistant, along with detailed CXR information -- including the radiology report, abnormality labels, bounding box coordinates, view, and gender (see prompt in \cref{supp_fig:conv_prompt}). Importantly, by leveraging the provided CXR information, the assistant is designed to respond as if it had direct visual access to the image (\cref{fig:conversation}). 

Following this process, we generate in total 89k image-conversation pairs, including 86k standard conversations (\cref{fig:conversation}a) and 3k ``grounded conversations'' (\cref{fig:conversation}b) that include interactions aimed at localizing specific observations. For the grounded conversations, it is essential to supply pairs of textual observations and their corresponding bounding box coordinates in the prompt. To achieve this, we use the datasets used for the ``phrase grounding'' task, MS-CXR and PadChest-GR, that provide annotations linking specific phrases in radiology reports to precise image location We thereby obtain 862 image-conversation pairs derived from MS-CXR grounded phrases, and 2,225 pairs derived from PadChest-GR (\cref{table:instruction-dataset-content}).

\subsection{Model finetuning}
\label{section:finetuning}

\begin{figure}[t]
    \centering
    \includegraphics[width=0.65\linewidth]{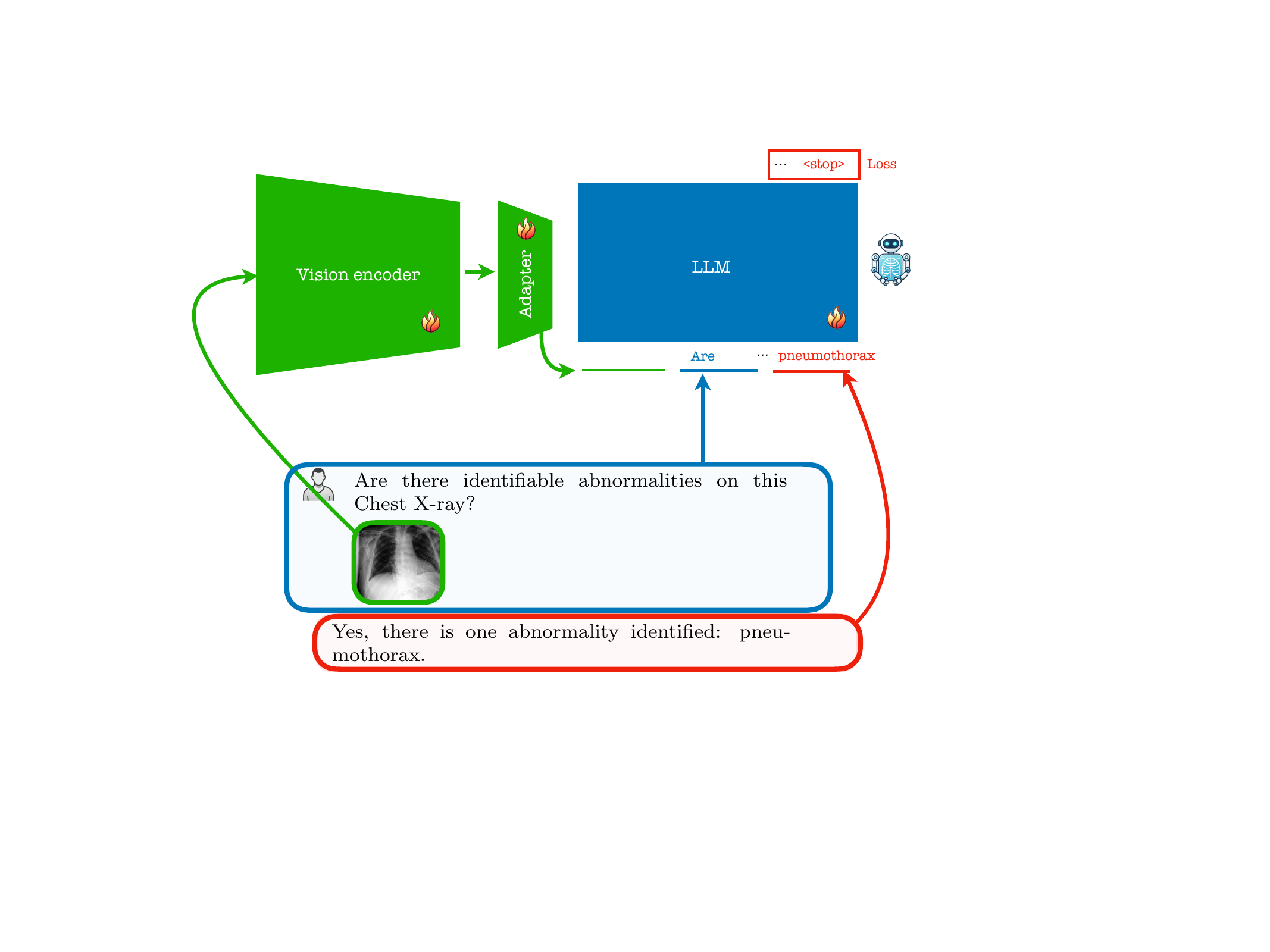} 
    \caption{ \textbf{Instruction fine-tuning of the vision-language model.} The CXR image is processed by the vision encoder, and the question is supplied at the language decoder. The flame icons indicate that the vision encoder, adapter, and LLM are all jointly fine-tuned end-to-end, generating the answer through next-token prediction.}
    \label{fig:finetuning} 
\end{figure}

We leverage an existing vision-language backbone, LLaVA-OneVision-7B \citep{li2024llava}. Its architecture is based on the SigLIP vision encoder \citep{zhai2023sigmoid} connected to the language model qwen-2 \citep{yang2024qwen2} via a 2-layer multi-layer perceptron (MLP). It was originally pretrained and instruction-tuned on image-text datasets in the general domain. We also follow the Higher AnyRes strategy \citep{chai2022any} by encoding multiple patches of the input image in different resolutions (in addition to the full image) and feeding the concatenated output representation from the vision encoder to the language model. \citet{chai2022any} showed that the scaling in image resolution is successful in the general domain. 

We fine-tune the LLaVA-OneVision-7B architecture on our instruction dataset by optimizing an auto-regressive loss on the target assistant tokens, $\mathbf{x}_{a}$. Specifically, for each token $x_i$ in the assistant's output sequence, we model
\begin{align}
    p(\mathbf{x}_a \mid \mathbf{x}_v, \mathbf{x}_q) 
    &= \prod_{i=1}^{L} p(x_i \mid \mathbf{x}_v, \mathbf{x}_{q,<i}, \mathbf{x}_{a,<i})~, 
\label{eqn:next-word-prediction}
\end{align}
where $\mathbf{x}_v$ denotes the visual tokens, and $\mathbf{x}_{q,<i}$ and $\mathbf{x}_{a,<i}$ represent the question and assistant tokens preceding token $x_i$, respectively. Here, $L$ is the number of tokens in the target assistant sequence.
This formulation also applies for the multi-turn conversations, where $\mathbf{x}_{q,<i}$ and $\mathbf{x}_{a,<i}$ include the chat history from previous rounds. Following recent trends in visual instruction tuning \citep{li2024llava, laurencon2024building}, the whole architecture is trained, using a learning rate of $2\text{e-}6$ in the vision encoder weights and $1\text{e-}5$ in the 2-layer MLP and language model weights. RadVLM is trained over 1 epoch of the instruction dataset using full-fine-tuning. We use 128 GH GPUs, each with 96GB of memory \citep{fusco2024understanding} for approximately 12 hours.  

For the ablation studies, we follow the same fine-tuning procedure but limit the process to a subset of the instruction dataset, focusing on isolated types of tasks.

%


\newpage 

\section{Experiments \& Results}

In this section, we describe our evaluation pipeline, the existing baseline models we use for comparison,  report results and highlight the capabilities of our RadVLM system.

\subsection{Evaluation pipeline}

In order to assess the quantitative performance of RadVLM, we design an evaluation pipeline based on the individual tasks from our instruction dataset. This pipeline leverages existing metrics for report generation, abnormality classification and visual grounding and creates novel evaluation tasks to assess the model's performance in conversational abilities. 

\subsubsection{Report generation}

To evaluate generated reports, we employ both lexical and radiology-specific metrics. Lexical metrics particularly quantify word overlap between generated and ground truth reports, among which we report BertScore, a metric for text generation based on computing token similarity using contextual
embeddings \citep{zhang2020BERTScore}, and Rouge-L \citep{lin2004rouge}, which quantifies the length of the longest common subsequence between predicted and reference reports. 

In contrast with lexical metrics, radiology-specific metrics ignore irrelevant variations in phrasing and focus on the clinically relevant semantics of the generated text, such as the presence or absence of an abnormality. In particular, we provide results for the RadGraph F1 and GREEN metrics.

\noindent
\textbf{RadGraph F1} \citep{delbrouck2022improving, yu2023evaluating}: computing this metric requires to map each generated and ground truth reports to a structured graph, named RadGraph \citep{jain2021radgraph}, containing radiology-specific entities (anatomy or observations) and the relations between them (``suggestive of'', ``located at''). The RadGraph F1 score \citep{delbrouck2022improving} computes the overlap in inferred structured graphs (extracted entities and relations) from both generated and ground truth reports. In our study, we use the recently released RadGraph-XL model to extract graphical representations \citep{delbrouck2024radgraph} and report the partial reward \citep[described in][]{delbrouck2022improving}. 

\noindent
\textbf{GREEN} \citep[Generative Radiology Report Evaluation and Error Notation,][]{ostmeier2024green}: a recently developed report generation metric that leverages the LLM-as-Judge mechanism of language models \citep{zheng2023judging} to identify clinically significant errors in generated radiology reports \citep{Yu2023-rt,calamida2023radiology,Calamida2024-le}, and that highly aligns with expert preferences as compared to other LLM-based evaluations (e.g., GPT-4).

\subsubsection{Abnormality classification}
We use the test split of the CheXpert dataset and prompt the model by asking to list the abnormalities present on the CXR. The mentioned abnormalities in the model's answer are extracted via key-word matching and compared to the ground truth list.
We calculate the F1 score for the 14 abnormalities and report the macro-averaged F1-score over all categories.

\subsubsection{Visual grounding}

We assess the model's performance in visual grounding by prompting the model to detect specific features (anatomy, abnormality, phrase) and extract the bounding box coordinates generated within the model's answer. Although we included the abnormality detection task in the instruction set, we do not evaluate it here due to the lack of comparable models for this task.

For the three grounding tasks, we use mean Average Precision (mAP) at an Intersection Over Union (IoU) threshold of 0.5. At this setting, we evaluate both precision and recall for predicted boxes that overlap with ground truth by at least 50\%.

\subsubsection{Multi-turn evaluation within conversational interactions}
\label{section:conversation-eval}

Evaluating conversational aspects of a model is essential to assess the utility and performance of an AI assistant in a multi-turn setting. We designed an LLM-based evaluation method carefully crafted for conversations, following the ``LLM-as-judge'' scheme \citep{zheng2023judging} previously adopted by LLaVA-Med \citep{li2023llava-med}. We created a test set of conversations following our generation process (see \cref{section:conversation}). As a result, we obtained  155 conversations containing grounding questions (derived from MS-CXR test set), and 500 conversations without (\cref{table:instruction-dataset-content}).  

During the evaluation process, we provide the CXR image to the VLM and sequentially ask the questions from this dataset. After collecting the VLM's answer to each question, we prompt GPT-4o (text-only) with the CXR information (report, list of abnormalities, etc.), the expected answer per question (derived from the ground truth information), and the VLM-generated answer (see \cref{supp_fig:eval_conv_prompt} for the prompt). At the end of the prompt,  GPT-4o is asked to provide an overall score (from 0 to 10) for the quality and accuracy of generated answers as compared to ground truth. We report two scores: one based on the grounded conversations dataset and one based on the non-grounded (standard) one.

 \subsection{Baseline models}
 \label{section:baseline-models}

\begin{table}[h]
\centering
\footnotesize
\renewcommand{\arraystretch}{1.5} 
\begin{tabular}{p{2cm}C{1cm}C{1.7cm}C{2cm}C{2cm}C{2cm}C{2cm}}
\toprule
\multicolumn{1}{c}{\textbf{Model}} & \textbf{Size} & \textbf{CXR-instr.} & \textbf{Report} & \textbf{Classification} & \textbf{Grounding} & \textbf{Conversation} \\ \hline
LLaVA-OV   & 7B  & 600     & \cmark & \xmark & \xmark & \cmark \\
LLaVA-Med  & 7B  & 8,060     & \cmark & \xmark & \xmark & \cmark \\
RaDialog   & 7B  & 580,000   & \cmark & \cmark & \xmark & \cmark \\
CheXagent  & 3B  & 8,500,000 & \cmark & \cmark & \cmark & \xmark \\
MAIRA-2    & 13B & 510,848   & \cmark & \xmark & \cmark & \xmark \\ 
\textbf{RadVLM} & 7B  & 1,052,162 & \cmark & \cmark & \cmark & \cmark \\ 
\bottomrule
\end{tabular}
\caption{\textbf{Functional capabilities and dataset sizes for RadVLM and baseline models.} Comparison of RadVLM with baseline models in terms of parameter size, number of unique CXR–instruction pairs used in training, and task coverage across report generation, abnormality classification, visual grounding, and conversational interactions.}
\label{table:baseline-models}
\end{table}

To provide insights into the performance of RadVLM in multiple tasks, we compare the instruction-tuned RadVLM to existing and publicly available state-of-the-art VLMs (from 3B to 13B, \cref{table:baseline-models}), adjusting the evaluation so that each baseline model is only tested on tasks for which it was initially trained on. Notably, in the general domain, we evaluate LLaVA-OneVision-7B \citep{li2024llava}, also used as our training starting point (\cref{section:finetuning}). In the medical domain, we evaluate LLaVA-Med \citep{li2023llava-med}, trained on a large-scale biomedical dataset (including CXRs). For CXR-specific VLMs, we use CheXagent (trained on a wide range of tasks), RaDialog (possessing conversation skills), and MAIRA-2 \citep{Bannur2024-ek}, specialized in report generation with visual grounding.

\subsection{Results}
\subsubsection{Report generation}

\begin{table}[t]
\centering
\footnotesize
\renewcommand{\arraystretch}{1.5} 
\begin{tabular}{l>{\centering\arraybackslash}p{1.8cm}>{\centering\arraybackslash}p{1.8cm}>{\centering\arraybackslash}p{1.8cm}>{\centering\arraybackslash}p{1.8cm}}
\toprule
 & \multicolumn{2}{c}{\textbf{NLG Metrics (\%)}} & \multicolumn{2}{c}{\textbf{Clinical Metrics (\%)}} \\ \cline{2-5}
 & BertScore & Rouge-L & RadGraph F1 & GREEN \\ \midrule
LLaVA-OV    & 32.9 & 13.5 & 6.6  & 13.1\\
LLaVA-Med   & 26.9 & 12.1 & 4.2  & 4.0 \\
RaDialog    & \underline{49.1} & 22.1 & 15.9 & 24.9 \\
CheXagent   & 37.4 & \underline{22.5} & \textbf{20.1} & \textbf{29.9} \\
MAIRA-2     & 46.6 & 17.7 & 12.9 & 21.3 \\ \midrule
RadVLM      & \textbf{51.9} & \textbf{25.4} & \underline{18.2} & \underline{27.7} \\ \bottomrule
\end{tabular}
\caption{\textbf{Performance of RadVLM and baseline models on single-image report generation.} NLG (Natural Language Generation) and clinical metrics are calculated on the filtered test set of MIMIC-CXR. Bold values indicate the best performance, while underlined values represent the second-best performance across models.}
\label{table:report-generation}
\end{table}

Firstly, we evaluate RadVLM on a core radiology task: generating a textual report from a frontal CXR. As our evaluation set, we use a filtered version of the MIMIC-CXR test set, excluding statements about findings from prior examinations. We report in \cref{table:report-generation} both lexical (BertScore, Rouge-L) and clinical (RadGraph F1, GREEN) metrics, which capture different performance aspects of the report generation task.

As mentioned in \cref{section:baseline-models}, five additional VLMs -- including three specific to CXR (RaDialog, CheXagent, MAIRA-2) -- are also evaluated, with minor adaptations to each model's recommended prompt template (\cref{supptable:prompts-comparison}). Many of these models were originally evaluated under different conditions (sometimes benefiting from extra inputs like prior images/reports or patient details or omitting certain metrics such as GREEN). To ensure that our setup remains consistent across all of them, we apply the same evaluation pipeline (test set and metrics).

For non-CXR specific VLMs, we observe a poor performance in both lexical and clinical metrics, with an unexpected improved performance for the generalist model (LLaVA-OneVision) over the medical one (LLaVA-Med), presumably profiting from a better architecture and training process \citep{li2024llava}. CXR-specific models perform significantly better than generalists, with a notable advantage of CheXagent in terms of clinical metrics. Although MAIRA-2 was reported to excel in report generation \citep{Bannur2024-ek}, we note that it was trained using additional inputs (indications, prior images, lateral views) that are not provided in our single‐image setup; thus, its lower performance here does not fully reflect its full capabilities. Overall, RadVLM achieves competitive results, attaining the highest performance in lexical metrics and the second-best in clinical metrics. This validates our approach, even though our training methodology was not specifically designed for the report generation task.

\subsubsection{Abnormality classification}

\begin{figure}[t]
    \centering
    \includegraphics[width=1.0\linewidth]{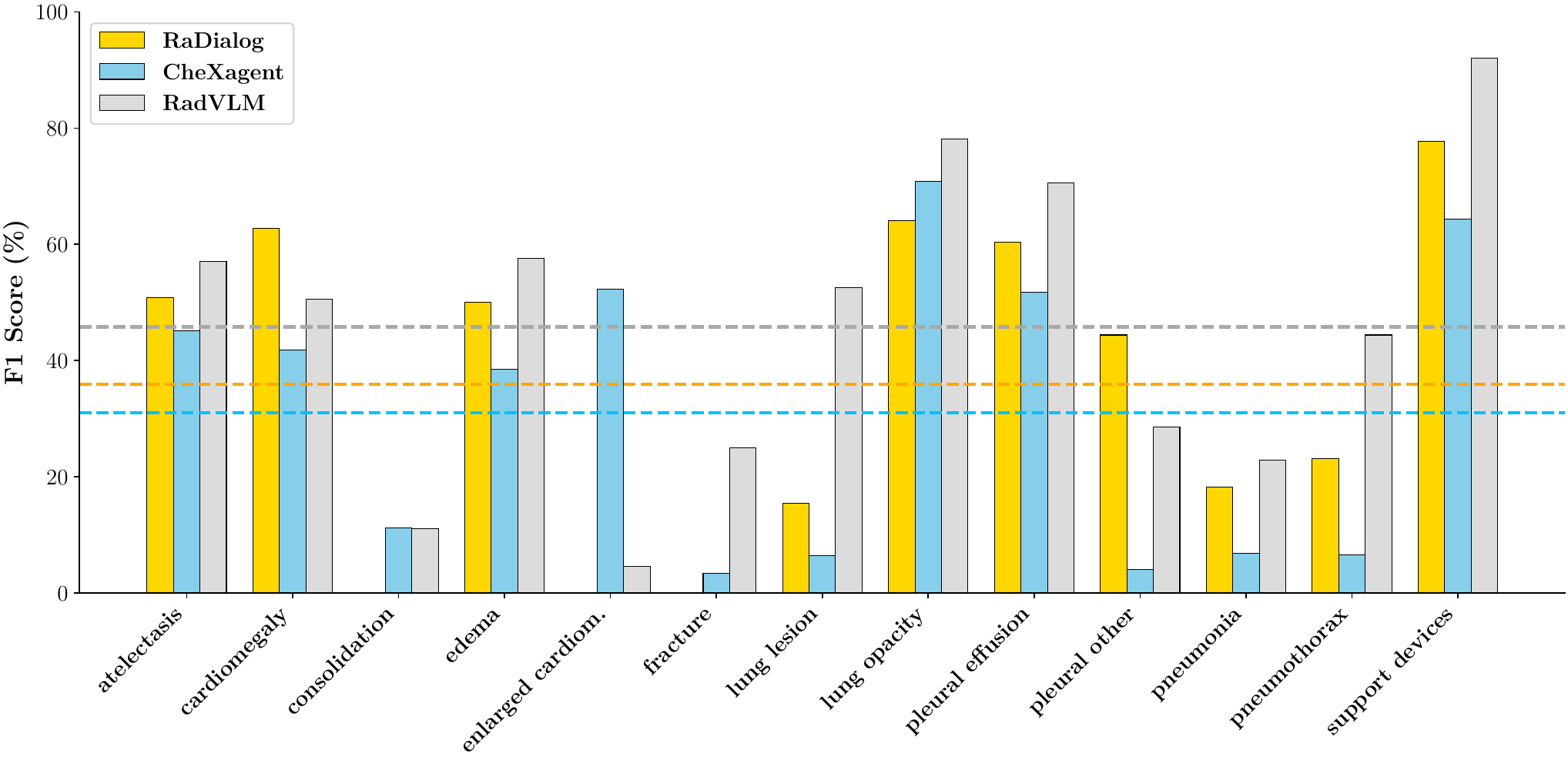} 
    \vspace{-5mm}
    \caption{\textbf{F1 scores for abnormality classification across different models.} Classification performance of \textcolor{gray}{\textbf{RadVLM (grey)}}, \textcolor{dandelion}{\textbf{RaDialog (yellow)}}, and \textcolor{cornflowerblue}{\textbf{CheXagent (blue)}}. Bars represent the F1 scores for individual pathology categories, while dashed lines indicate the macro-averaged F1 score across all categories. Note that RaDialog was trained exclusively on MIMIC-CXR labels and therefore evaluated on out-of-domain data.}
    \label{fig:classif} 
\end{figure}

In this section, we assess each model's ability to predict which abnormalities are visible on the CXR.  We use the manually curated CheXpert test set, prompt RadVLM to list any observed abnormalities, and compare them against the ground truth. As for the report generation task, we adapt the prompt for each compared baseline model (RaDialog, CheXagent) so they output a full list of abnormalities. Notably, RaDialog can identify pathologies from a single open-ended question, whereas CheXagent needs all possible labels explicitly listed in its prompt (see \cref{supptable:prompts-comparison} for details).
 
\cref{fig:classif} shows the individual F1 scores for the 14 CheXpert labels as well as the macro-averaged F1 across all categories (dashed lines in \cref{fig:classif}). Our findings suggest that RadVLM better captures which abnormalities appear on the CXR, reflected in its higher macro-F1 score. In particular, it shows improved classification for crucial pathologies such as atelectasis, edema, fracture, lung lesion, lung opacity, pneumonia, pleural effusion, and pneumothorax. While CheXagent showed strong classification results under binary or restricted-label evaluations \citep{chen2024chexagent}, it struggles when tasked with identifying all categories at once (see \cref{fig:classif}). We also note that RaDialog -- having been trained exclusively on MIMIC-CXR labels -- was evaluated on entirely out-of-domain data, which likely impacted its overall classification results. 
 
Overall, these results indicate that RadVLM performs equally or better than its counterparts in abnormality classification. This step is especially crucial for the grounding process since identifying the relevant pathology is a prerequisite to localizing it on the CXR.

\subsubsection{Visual grounding}

In this section, we evaluate RadVLM's visual grounding capabilities, which could help clinicians localize specific regions or pathologies on a CXR. This is particularly useful once a pathology has already been identified -- either by a radiologist's input or through our previously described AI tasks -- since it allows one to pinpoint exactly where the abnormality appears on the image.

We report performance metrics for the three main grounding tasks RadVLM was trained on: anatomical grounding using the Chest Imagenome test set, abnormality grounding using the VinDr-CXR test set, and phrase grounding using the MS-CXR test set (\cref{table:instruction-dataset-content}). For each task, we use mean Average Precision (mAP) as our primary evaluation metric.

As mentioned in \cref{table:baseline-models}, some of the CXR-specific VLMs already have grounding capabilities. CheXagent was trained to handle both abnormality and phrase grounding tasks, while MAIRA-2 -- originally trained to produce radiology reports with grounded observations -- is also capable of predicting bounding box coordinates when provided with input text. After retrieving each model's instruction template for generating bounding box coordinates (\cref{supptable:prompts-comparison}), we evaluated both CheXagent and MAIRA-2 on all three grounding tasks performed by RadVLM.

\begin{figure}[t]
\centering
{\footnotesize
\begin{minipage}[t]{0.48\linewidth}
  \vspace{0pt}
  \textbf{a. Anatomical grounding}
  \human{Where is the [region] located on the image?
    \vspace{1mm}
  }
  \assistant{The [region] is located at \textcolor{red}{[coordinates]} on the image. 
  \vspace{2mm}
  
    \begin{minipage}{\linewidth}
      \centering
      \regionimage{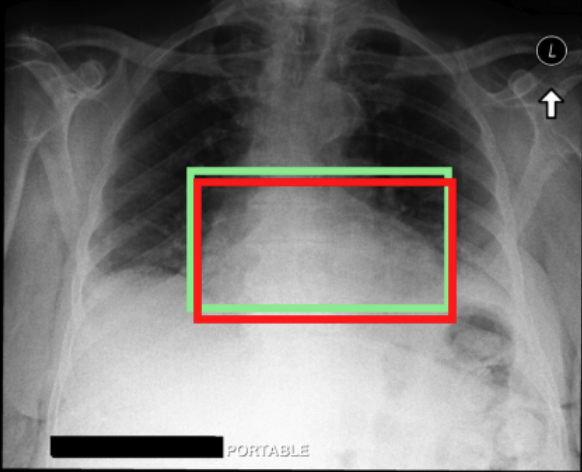}{cardiac \\[-1mm] silhouette }\hfill
      \regionimage{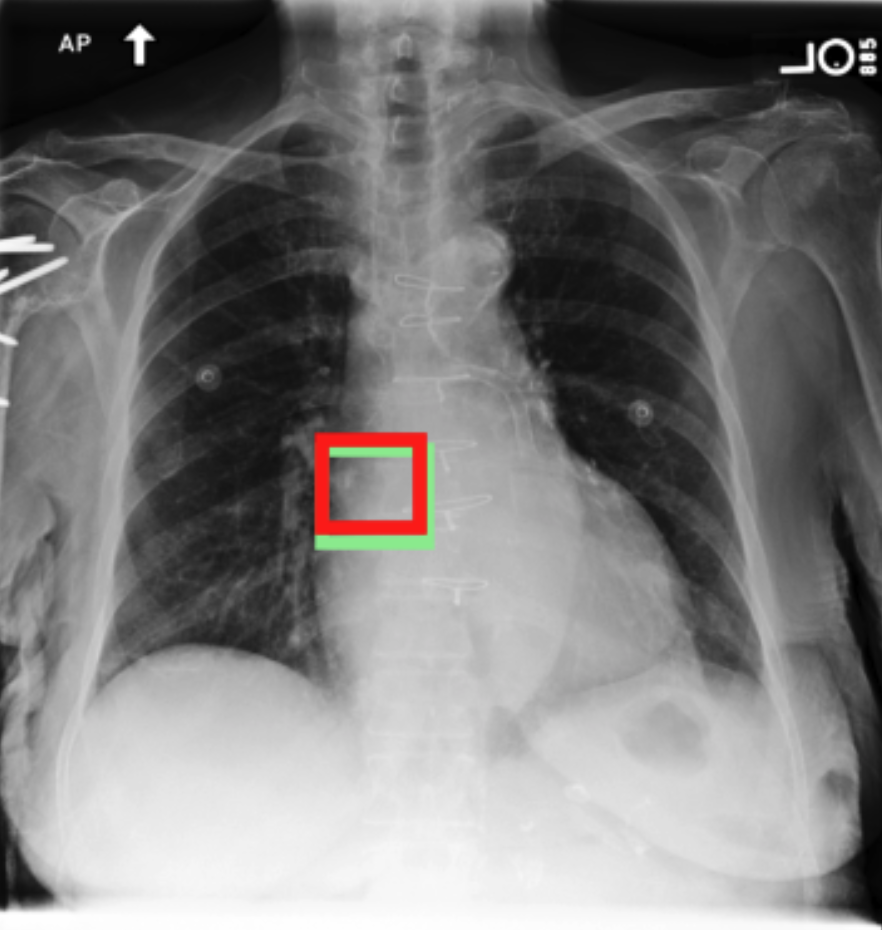}{cavoatrial \\[-1mm] junction}\hfill
      \regionimage{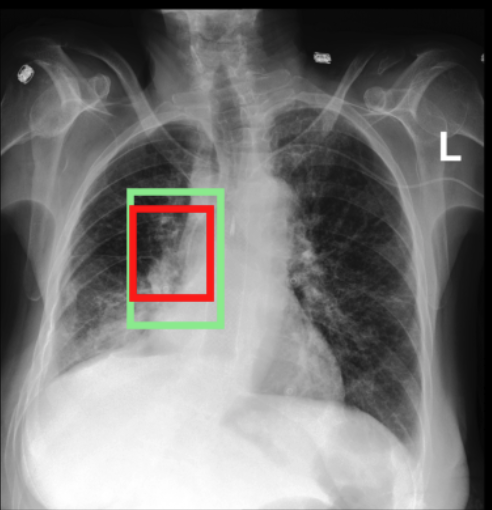}{right hilar \\[-1mm] structures}\hfill
      \regionimage{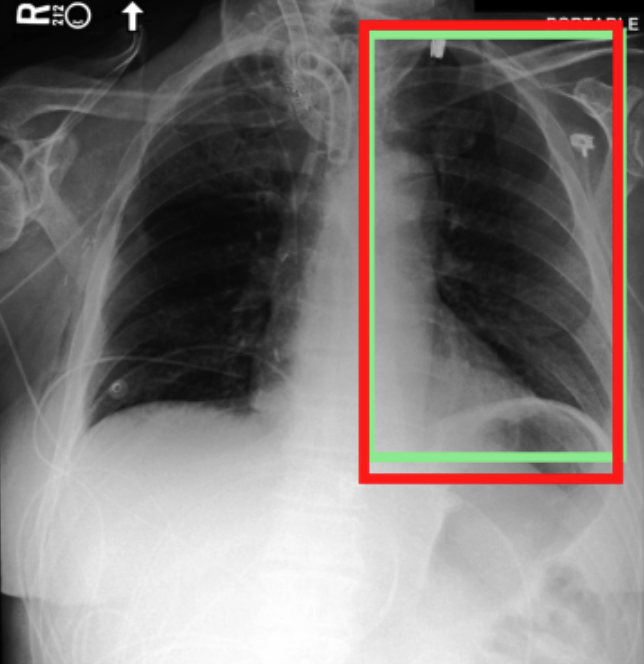}{left\\[-1mm] lung}
      \\ \vspace{1mm} 
      \regionimage{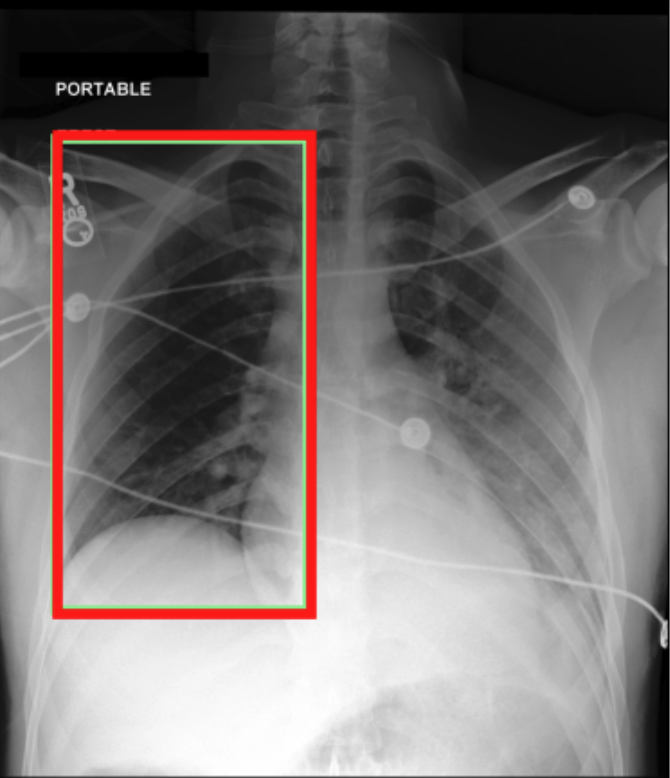}{right lung}\hfill
      \regionimage{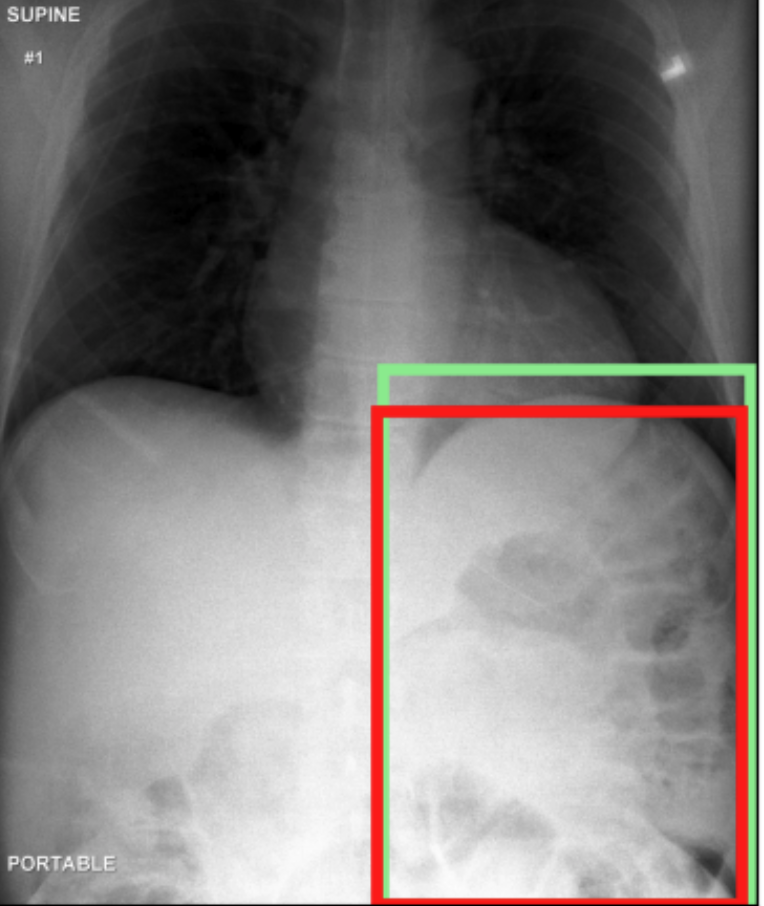}{left upper \\[-1mm] abdomen}\hfill
      \regionimage{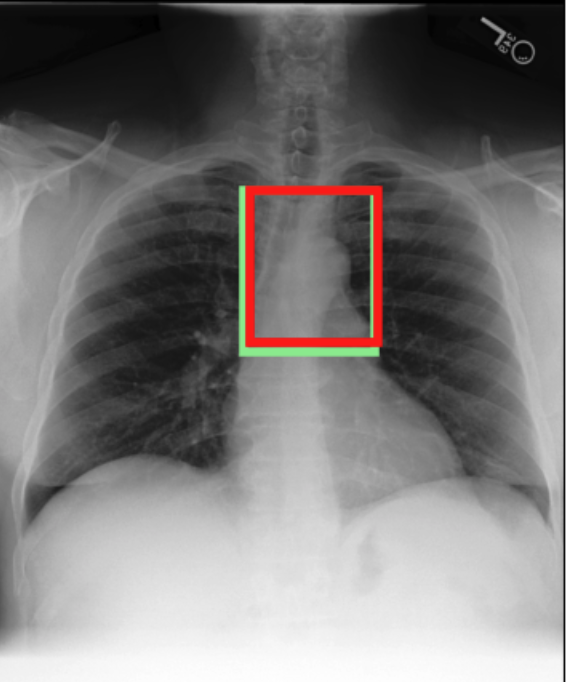}{upper \\[-1mm] mediastinum}\hfill
      \regionimage{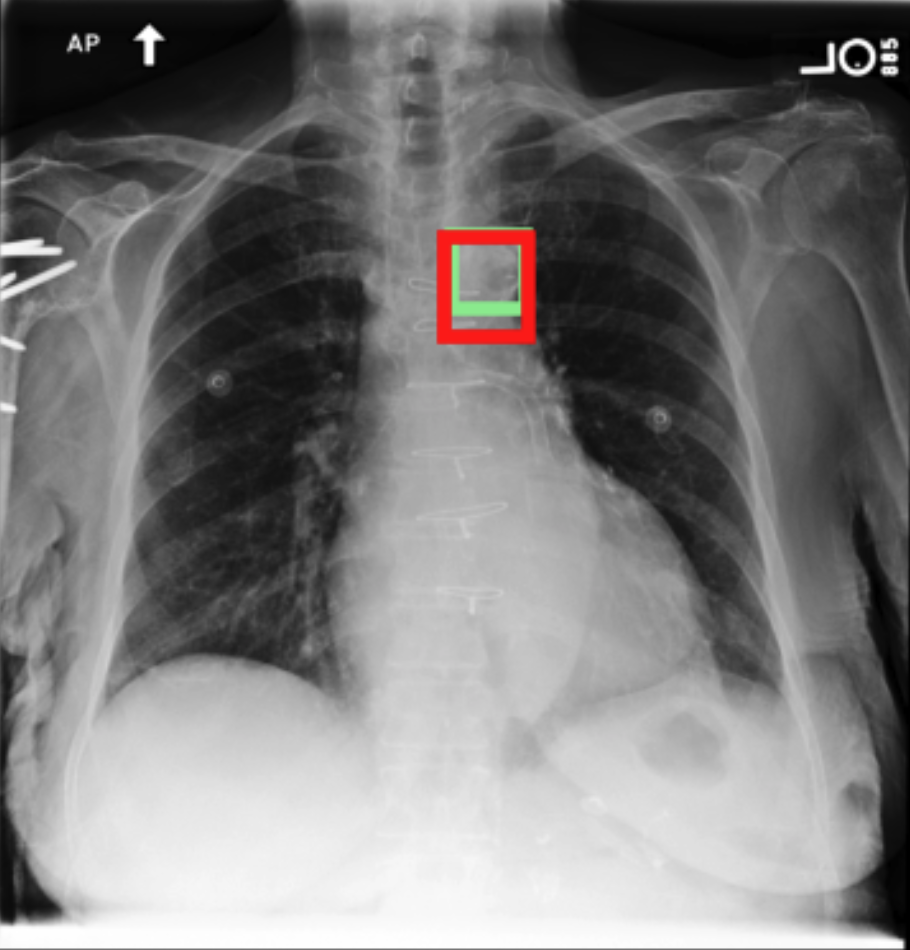}{aortic \\[-1mm] arch}
    \end{minipage}
    \vspace{2mm}
  }
  \centering
\end{minipage}%
\hfill
\begin{minipage}[t]{0.48\linewidth}
  \vspace{0pt}
  \textbf{b. Abnormality grounding}
  \human{Locate the [abnormality] on the CXR.
   \vspace{1mm}
  }
  \assistant{The [abnormality] is positioned at the coordinates \textcolor{red}{[coordinates]} on the CXR.
  \vspace{2mm}
  
    \begin{minipage}{\linewidth}
      \centering
      \regionimage{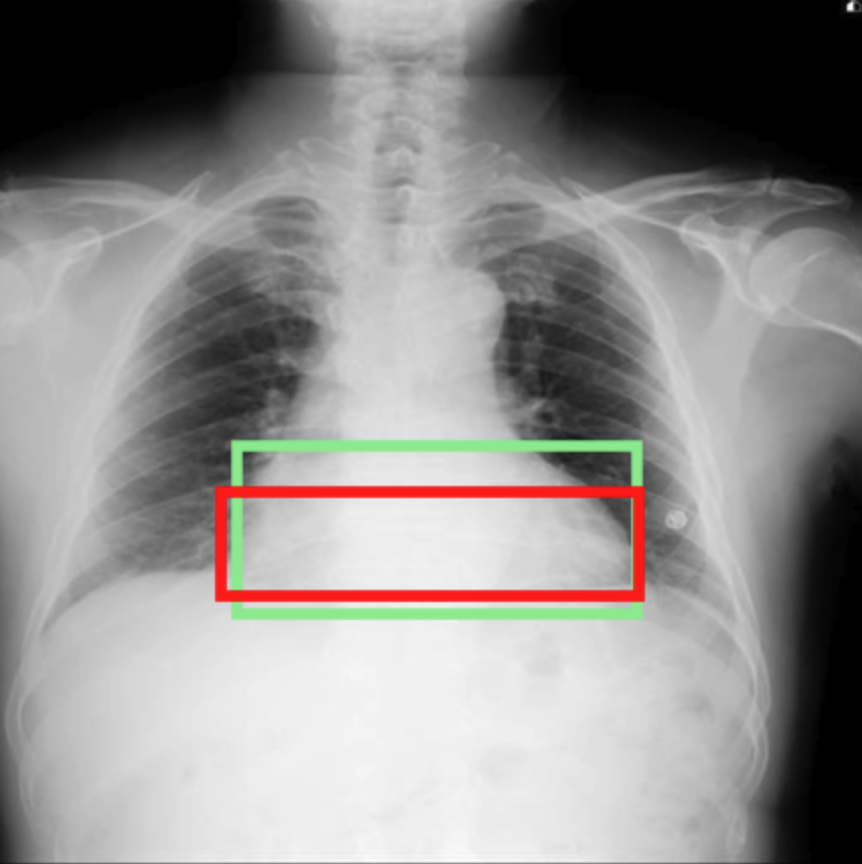}{cardiomegaly}\hfill
      \regionimage{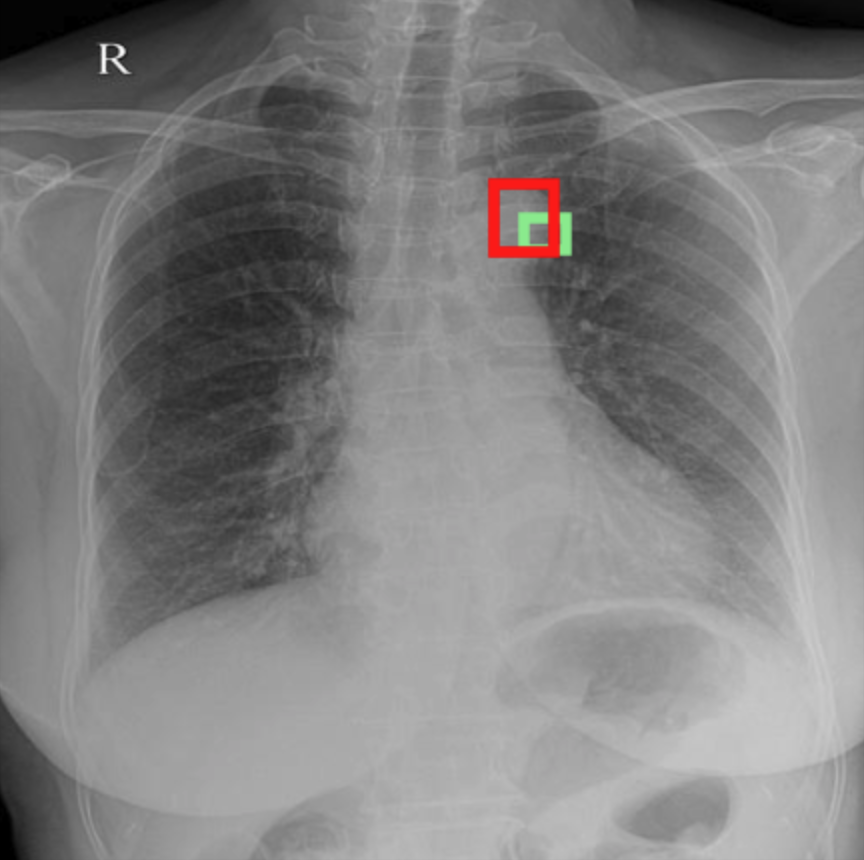}{calcification}\hfill
      \regionimage{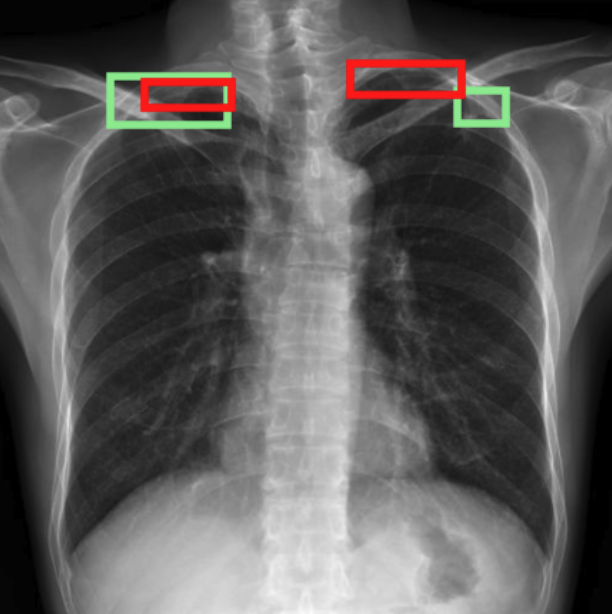}{pleural \\[-1mm] thickening}\hfill
      \regionimage{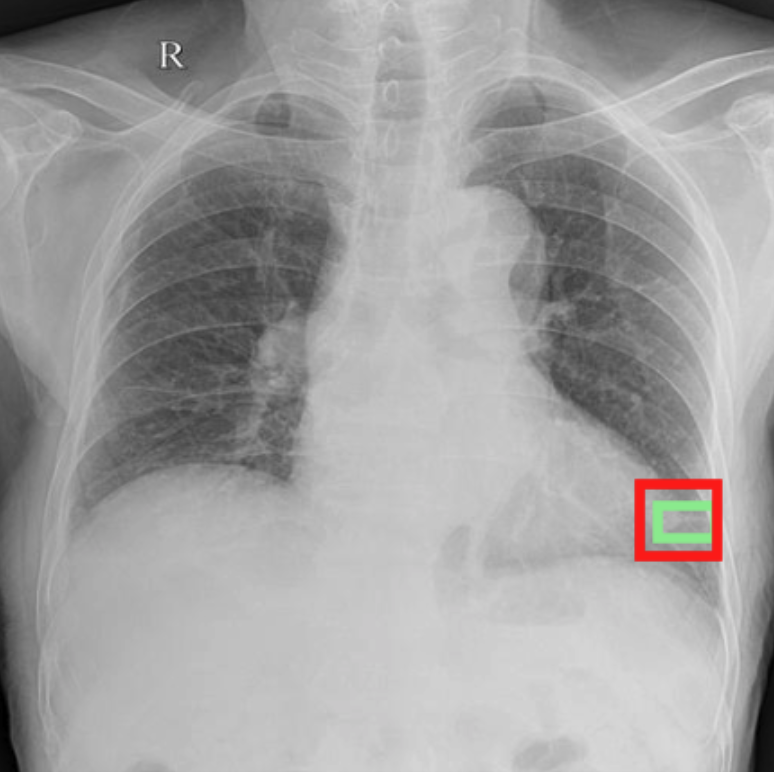}{atelectasis}
      \\[1mm] 
      \regionimage{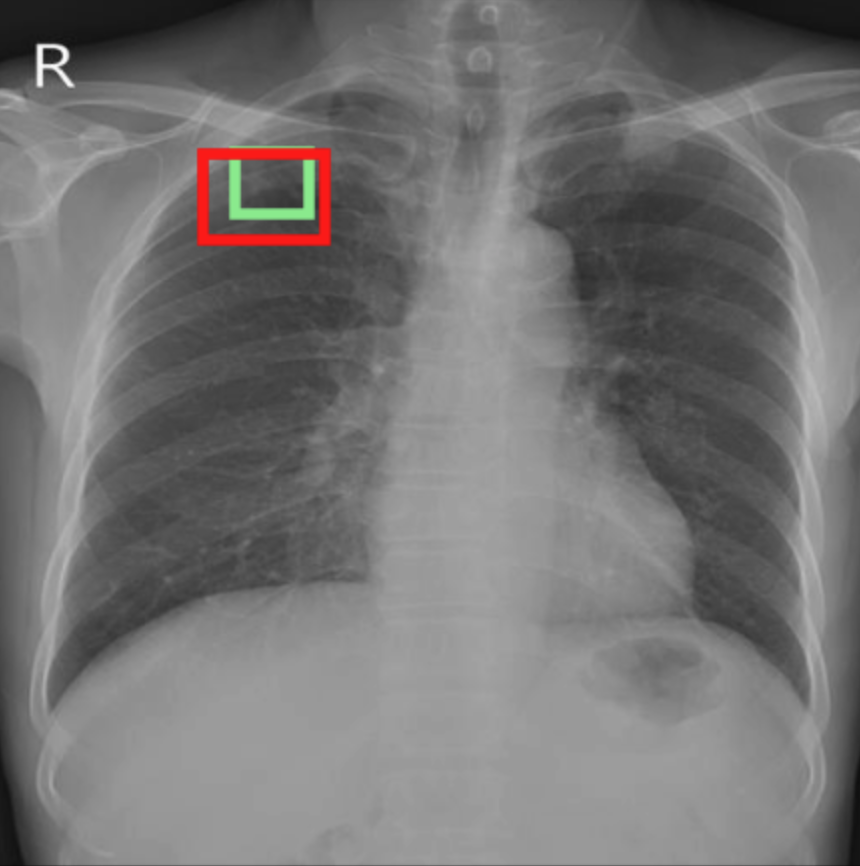}{pulmonary \\[-1mm] fibrosis}\hfill
      \regionimage{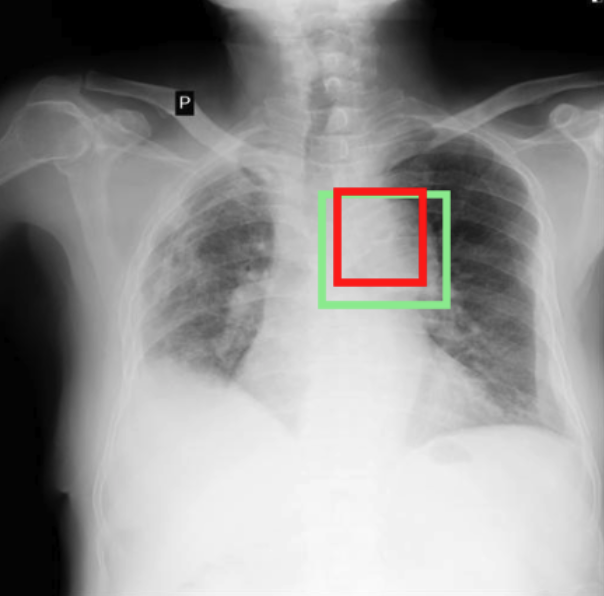}{aortic \\[-1mm] enlargement}\hfill
      \regionimage{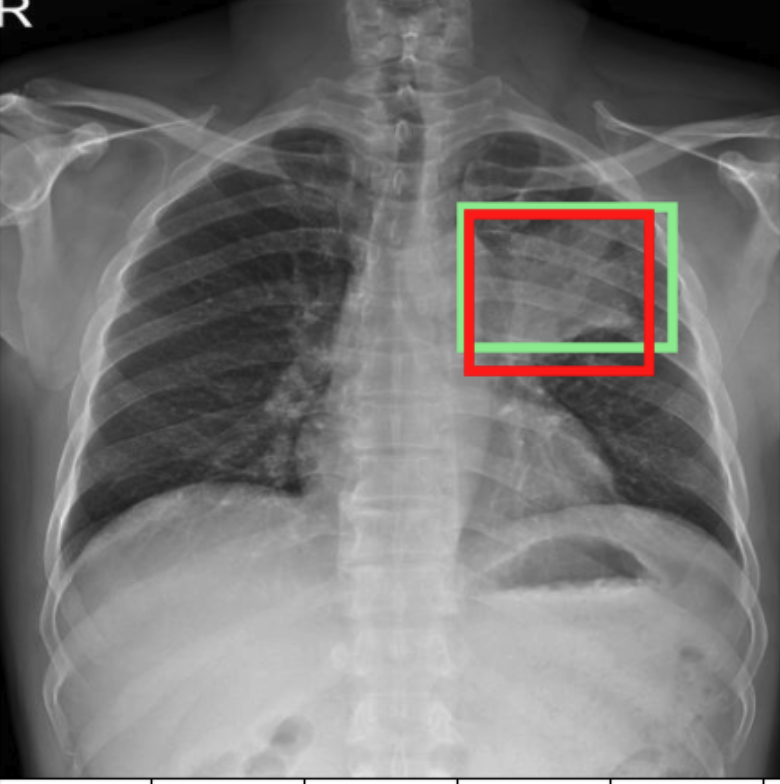}{consolidation}\hfill
      \regionimage{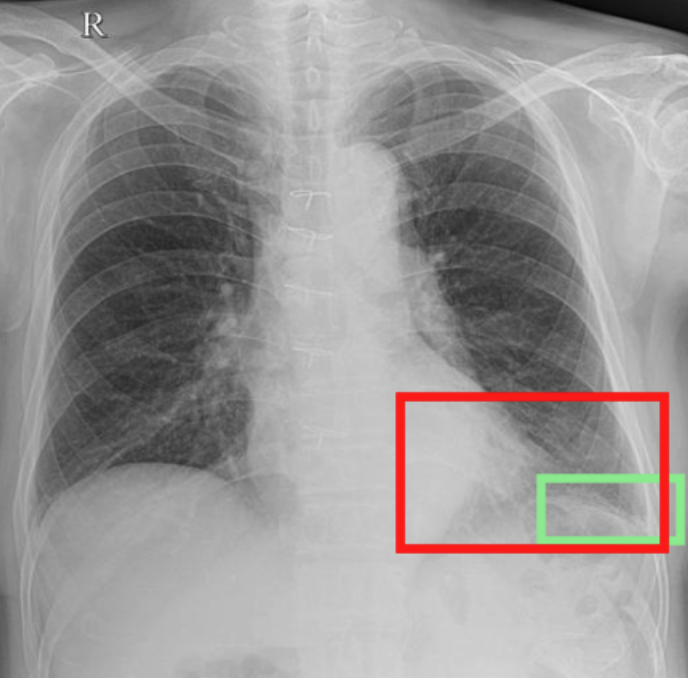}{interstitial \\[-1mm] lung disease}
    \end{minipage}
    \vspace{2mm}
  }
  \centering
\end{minipage}
}
\vspace{4mm}
\caption{\textbf{CXR region and abnormality grounding with RadVLM.} Examples of RadVLM's grounding predictions for \textbf{(a) anatomical regions} and \textbf{(b) abnormalities} in CXR images. The model predicts bounding boxes indicating the location of queried structures or pathological findings. Green boxes represent ground truth annotations, while red boxes denote model-predicted bounding boxes.}
    \label{fig:grounding-results} 
\end{figure}

Our results show that RadVLM performs well at localizing anatomical regions (e.g., ``right lung'', ``aortic arch'', illustrated in \cref{fig:grounding-results}a), achieving a mAP of 85.8 \%, by far surpassing the other CXR grounding models (\cref{table:grounding-results}). This advantage is partly explained by including the Chest Imagenome dataset (and thus the anatomical grounding task) in the training set, which CheXagent and MAIRA-2 did not leverage. However, it remains a key feature for any grounding model to possess a fine-grained understanding of CXR anatomy. 

For the abnormality grounding task, RadVLM is less consistent (\cref{fig:grounding-results}b), likely due to higher sparsity of abnormality locations and labels, yet it still achieves best performance (\cref{table:grounding-results}).  For the phrase grounding task, while MAIRA-2 and CheXagent demonstrate great performance, RadVLM surpasses them with a mAP of 81.8\% (\cref{table:grounding-results}), presumably benefiting from the newly released PadChest-GR dataset \citep{castro2024padchest} used for training. 

Overall, these results show that our instruction tuning strategy for visual grounding (covering three essential tasks), combined to a modern VLM backbone, offers a promising avenue to help clinicians localize anatomical and pathological features during a CXR exam. Furthermore, providing fine-grained details within an LLM-generated output may also enhance the ability to answer grounded questions in a multi-turn setting, as we explore next.

\begin{table}[t]
\centering
\footnotesize
\renewcommand{\arraystretch}{1.5} 
\begin{tabular}{p{1.5cm}>{\centering\arraybackslash}p{2cm}>{\centering\arraybackslash}p{2cm}>{\centering\arraybackslash}p{2cm}}
\toprule
\multicolumn{1}{c}{} & \textbf{Anatomical grounding}  & \textbf{Abnormality grounding} & \textbf{Phrase grounding}  \\ \midrule
CheXagent & 6.2  & \underline{26.0} & 69.7 \\
MAIRA-2  & \underline{19.8} & 11.3 & \underline{80.1} \\ \midrule
\textbf{RadVLM} & \textbf{85.8} & \textbf{34.6} & \textbf{81.8}\\ 
\bottomrule
\end{tabular}
\caption{\textbf{Visual grounding performance measured in mAP (\%)}. Mean average precision (mAP) scores for anatomical grounding, abnormality grounding, and phrase grounding tasks across different models. Bounding box coordinates are extracted from each model's output, and mAP is computed using an IoU threshold of 50\%. \textbf{Bold} values indicate the highest performance, while \underline{underlined} values represent the second-best performance.}
\label{table:grounding-results}
\end{table}

\subsubsection{Conversational abilities}

Our model can already handle multiple advanced tasks for CXR interpretation. However, we have only tested its ability to follow instructions in a single-turn, question-answer format so far. An essential aspect of a VLM's performance is also its capacity to manage follow-up questions in a multi-turn exchange with the user. Simply generating a full report or marking image features in one shot does not necessarily imply that the model can handle a varied sequence of precise questions -- covering observations, locations, clarifications of medical terms, and more.

RadVLM's instruction dataset also includes image-conversation pairs (\cref{fig:conversation}) generated with GPT-4o, by providing it with a radiology report, any available grounded phrases, and additional details. As mentioned in \cref{section:conversation-eval}, we created two held-out evaluation sets using GPT-4o: one without grounded questions and one focused on grounded questions derived from the MS-CXR test set (\cref{table:instruction-dataset-content}). We run the evaluation on both test sets and report the average score across all samples. We also evaluate the other baseline models that possess conversational abilities, thoroughly respecting their prompting template within a multi-turn setting. The average GPT-4o scores for RadVLM and other models are gathered in \cref{table:eval_conv}.

\begin{figure}[h!]
\footnotesize 
  \human{What does the chest X-ray show?
  \vspace{1mm} 
  
  \roundimage{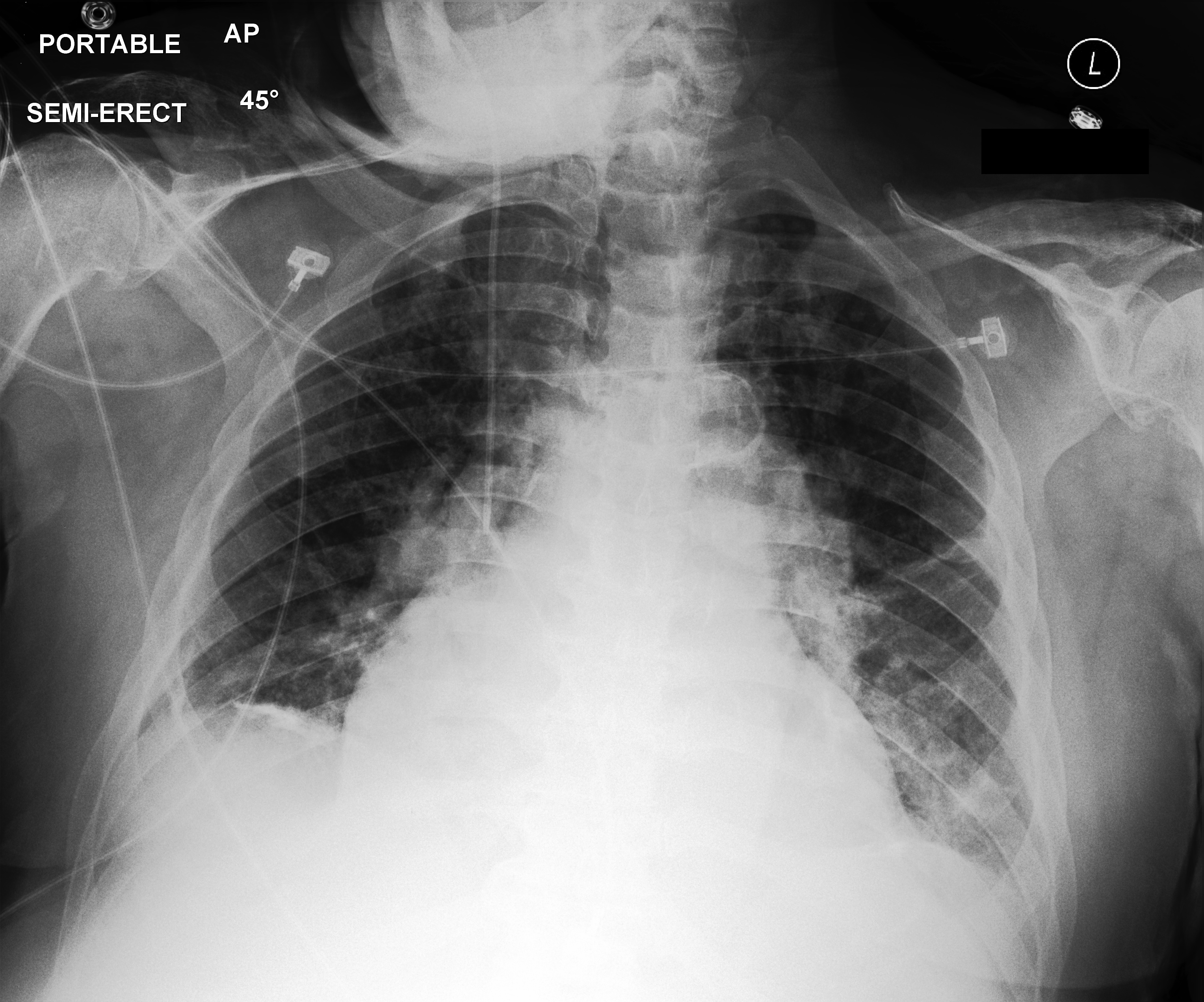}{0.1}
  }
  \reference{The X-ray shows multiple important findings, including indications of pulmonary edema, cardiomegaly (an enlarged heart), and some scarring on the left side of the lung. It also shows a catheter in place.}
  \assistant{\textbf{The X-ray shows some notable findings, including an enlarged heart size and areas of increased density in the lungs, which could suggest fluid buildup or other issues. There are also some medical devices visible.}}
  \radialog{The patient has cardiomegaly, lung opacity, atelectasis, pleural effusion and support devices.}
  \llavamed{ The chest X-ray shows bilateral infiltrates, which are abnormal substances or collections of cells that have entered the lung tissue. These infiltrates can be seen in both lungs and may indicate an underlying condition, such as an infection, inflammation, or other lung-related issues.}
  \human{Is the heart size normal on this X-ray?}
  \reference{The heart appears enlarged, a condition called cardiomegaly.}
    \assistant{\textbf{The heart appears larger than normal, a condition referred to as cardiomegaly. This can sometimes indicate underlying heart-related issues, but further evaluation would be needed to understand the cause.}}
    \radialog{No, the patient has cardiomegaly.}
    \llavamed{Yes, the heart size appears to be normal in this chest X-ray.}
    
  \human{Can you tell me where the heart enlargement is visible?}
  \reference{The enlargement of the heart can be seen within the area defined by the coordinates [0.27, 0.59, 0.77, 0.93] on the X-ray.}
  \assistant{\textbf{The enlarged heart is visible within the bounding box coordinates [0.26, 0.53, 0.79, 0.91] on the image.}}
  \radialog{The heart enlargement is visible in the left hilum.}
  \llavamed{The heart enlargement is visible in the chest X-ray taken on the 1st day of admission. However, it is not visible in the image you are referring to, which is the chest X-ray taken on the 3rd day of admission.}
  \human{ Does the X-ray suggest any issue with air in the chest cavity?}
  \reference{No, the X-ray does not show any evidence of pneumothorax, which would be the presence of air in the chest cavity outside the lungs.}
  \assistant{\textbf{No, there is no evidence of air outside the lungs, which would indicate a pneumothorax. The chest cavity appears normal in that regard.}}
  \radialog{Yes, the patient has air in the chest cavity.}
  \llavamed{Yes, the X-ray suggests the presence of air in the chest cavity, which is referred to as pneumomediastinum (...)}
  \human{What kind of X-ray view is this?}
  \reference{This X-ray was taken in the AP (anteroposterior) view.}
  \assistant{\textbf{This is an AP (anteroposterior) view.}}
  \radialog{This is a lateral view.}
  \llavamed{This is a posteroanterior (PA) view of a chest X-ray.}
  
\vspace{5mm}

\caption{\textbf{Example of multi-turn interaction with RadVLM and other conversational VLMs.} Comparison of expected answers from the \textcolor{pastelred}{\textbf{conversation dataset} (red bubbles)} to generated answers from \textcolor{gray}{\textbf{RadVLM} (grey bubble)}, \textcolor{dandelion}{\textbf{RaDialog} (yellow bubble)}, and  \textcolor{pastelpurple}{\textbf{LLaVA-Med} (purple bubble)}.}
\label{fig:eval-conversation}
\end{figure}

Our results show that RadVLM, trained on a broad range of image-conversation pairs, achieves an average score of 6.66/10, substantially higher than the other conversational models, whose responses often prove incorrect or vague (\cref{fig:eval-conversation}). This advantage may stem from RadVLM's ability to handle varied question types and sequences in multi-turn exchanges, a skill reinforced by the sparse nature of its training conversations. The gap becomes even more pronounced in grounded scenarios, where RadVLM maintains strong performance (6.60/10) while others drop even more. Notably, this suggests that even the limited number of grounded conversations included (\cref{table:instruction-dataset-content}) was sufficient to equip RadVLM with robust grounding capabilities in a multi-turn setting.

\begin{table}[t]
\centering
\footnotesize
\renewcommand{\arraystretch}{1.5} 
\begin{tabular}{>{\centering\arraybackslash}p{2.8cm} >{\centering\arraybackslash}p{2.3cm} >{\centering\arraybackslash}p{2.3cm}}
\toprule
\multicolumn{1}{c}{\textbf{Model}} & \textbf{GPT-4o Score} & \textbf{GPT-4o Score} w/ grounding \\ \midrule
\multicolumn{1}{l}{LLaVA-OV}   & 3.66 & 2.95 \\
\multicolumn{1}{l}{LLaVA-Med}  & 3.52 & 2.95  \\
\multicolumn{1}{l}{RaDialog}   & \underline{3.62} & \underline{3.11}  \\ \midrule
\multicolumn{1}{l}{\textbf{RadVLM}}  & \textbf{6.66} & \textbf{6.60} \\ 
\bottomrule
\end{tabular}
\caption{\textbf{Conversation performance on standard and grounded interactions.} GPT-4o evaluation scores for conversational performance across models, comparing generated responses to expected answers derived from image-based attributes (e.g., reports, bounding boxes, and view type). Scores are reported for standard conversations and grounded conversations, where models must provide spatially aware responses. \textbf{Bold} values indicate the best performance, while \underline{underlined} values represent the second-best performance.}
\label{table:eval_conv}
\end{table}

\subsection{Ablation studies}

In this section, we investigate a fundamental question in LLM fine-tuning: do models perform individual tasks more effectively when they are trained separately, or does joint training on multiple tasks provide a greater overall benefit? To address this question, we perform ablation studies where we train separate models -- each dedicated to a single task (e.g., report generation, classification) -- using the same training settings and hyperparameters as RadVLM, then evaluate their performance on that respective task.

Our results reveal that RadVLM generally outperforms models trained on individual tasks. First, this difference is pronounced in grounding performance, particularly for tasks with smaller training sets (e.g., phrase grounding; see \cref{table:instruction-dataset-content}).  We attribute this to the complementary nature of the various grounding tasks, which collectively improve the model's ability to localize visual elements (including overlapping categories, such as ``abnormalities'' also mentioned in ``phrases'') and prevent overfitting. 

A similar trend emerges in the conversation scores, where RadVLM outperforms a model trained solely on the conversation dataset. While this ablated model still performs significantly better than its baseline backbone -- surpassing LLaVA-OneVision's score in \cref{table:baseline-models} -- RadVLM leverages knowledge from the other tasks, such as report generation, that contains a higher amount of data points. This presumably enables it to generalize more effectively to new questions from the conversation test set. 

Report generation is the only task that turns out equivalent when trained in isolation. One likely reason is the high amount of training data for this task, and the richness of information textual reports contain over other tasks, which might suffice for the model to generalize over new instances.  

\begin{table}[h]
\centering
\footnotesize 
\renewcommand{\arraystretch}{1.5} 
\begin{tabular}{lcccccccc}
    \toprule
   \multirow{2}{*}{\vspace{-17mm}\textbf{Training tasks}} & \multicolumn{8}{c}{\vspace{1mm}\textbf{Evaluation tasks}} \\ 
   & \parbox{1.8cm}{\centering Report generation} 
   & \parbox{1.6cm}{\centering Classification} 
   & \parbox{1.6cm}{\centering Anatomical grounding} 
   & \parbox{1.6cm}{\centering Abnormality grounding} 
   & \parbox{1.6cm}{\centering Phrase grounding}
   & \parbox{1.8cm}{\centering Conversation (standard/ grounded)} \\ \cline{2-8}
   & RadGraph/GREEN & F1 (macro) (\%)&mAP (\%)&mAP (\%)&mAP (\%)&GPT-4o-Score \\ \midrule
Report generation       & 18.5 / 27.6  & -   & -   & -   & -   & -   \\  
Classification          & -       & 41.5 & -   & -   & -   & -    \\  
Anatomical grounding    & -       & -   & 83.0 & -   & -   & -    \\  
Abnormality grounding   & -       & -   & -   & 27.1 & -   & -    \\  
Phrase grounding        & -       & -   & -   & -   & 64.6 & -    \\  
Conversation            & -       & -   & -   & -   & -   & 6.11 / 5.89  \\  
All (RadVLM)                     & 18.2 / 27.7 & \textbf{45.8} & \textbf{85.8} & \textbf{34.6} & \textbf{81.8} & \textbf{6.66} / \textbf{6.60}  \\ \bottomrule
\end{tabular}
\caption{\textbf{Ablation studies comparing single-task models to RadVLM.}  Each ablated model is trained on only one task from RadVLM’s instruction dataset and evaluated on that same task. We compare these results to RadVLM’s previously reported performance over all tasks. }
\
\end{table}

\section{Conclusions}

There is a growing need for medical AI tools that are both accurate and user-friendly, and can interact with clinicians in a natural, multi-turn conversation. In this study, inspired by advances in both general and CXR-specific vision-language models, we explored the potential of constructing an assistant that not only aims for strong performance in individual tasks but also for handling them in a multi-turn, interactive setting. To achieve this, we built an instruction dataset that integrates both stand-alone tasks and multi-task dialogues, then fine-tuned a VLM backbone on it. Our findings suggest that this method performs well and offers promising insights for designing future radiology assistants. 

One key contribution of this study is our consistent re-implementation and evaluation of existing models (alongside RadVLM) under a unified experimental framework. In an era where AI research is proliferating rapidly, and new models are introduced almost daily, it can be challenging to compare results that rely on different setups. By systematically re-implementing models and replicating experiments within the same context, we enhance both reproducibility and reliability. This standardized approach aligns with initiatives such as the ReXrank leaderboard for CXR report generation \citep{zhang2024rexrank}, which similarly evaluates state-of-the-art models using a consistent protocol. Ultimately, these efforts provide a clearer understanding of each model's true capabilities and help researchers identify the most promising directions for future work. 

Another key takeaway is the flexibility of VLMs in handling diverse data types, ranging from free-text reports and classification labels to fine-grained information, all within a multi-turn, ``real-life'' scenario. Traditionally, each task might have been tackled by distinct, specialized architectures -- e.g., separate object detectors \citep{sun2022research} or classifiers \citep{he2016deep} -- but our findings support the recent trend, employed in both general and CXR-specific visual modeling  \citep{lin2024vila, wang2024qwen2vl, sharma2024maira, chen2024chexagent}, where many data formats can now be tokenized and processed through a common next-token generation paradigm. Our ablation studies further show that while each task differs in purpose, they collectively enhance the model's overall understanding of a CXR image, an effect likely amplified by training them together. Even though multi-agent approaches emerge in the medical setting \citep{schmidgall2024agentclinic}, these results support the value of joint training for developing comprehensive, single-agent solutions. 

The simplicity of RadVLM's training strategy also sheds light on potential directions for future CXR-based modeling. Here, we built upon LLaVA-OneVision \citep{li2024llava}, a VLM already pretrained for instruction following, and fine-tuned it on our curated dataset for just a single epoch. Although other studies have employed more intricate pipelines, such as pre-training the vision encoder, adapter and LLM components successively \citep{chaves2024llavarad, chen2024chexagent, Bannur2024-ek}, Given the competitive performance of RadVLM with other models, we hypothesize that the quality of the instruction dataset, coupled with the backbone architecture \citep[as shown by][]{laurenccon2024matters} are the most critical for the overall performance of such a visual foundation model. That said, for specific tasks like report generation, specialized solutions that incorporate additional patient details, prior and lateral images, or extended training, could offer greater gains. This likely explains why our model does not surpass the state of the art for that particular task. From this, we envision a future where AI assistants are bifurcated into \emph{(i)} non-interactive, report-focused tools designed for experienced radiologists, and \emph{(ii)} more adaptable, conversational agents that clinicians or radiology students can easily query in a multi-turn context. 

RadVLM serves as a strong starting checkpoint for a second optimisation stage based on modern reinforcement–learning (RL) techniques that directly reward reasoning quality and clinical accuracy.  
Recent language–model work such as DeepSeekMath and DeepSeek-R1 introduces Group Relative Policy Optimisation (GRPO), a critic-free policy-gradient method that improves overall answer accuracy by encouraging more effective reasoning traces \citep{shao2024deepseekmath,guo2025deepseek}. GRPO has since been extended to vision–language models in both general \citep{shen2025vlm} and medical domains \citep{lai2025med}, showing consistent gains in localization and diagnostic tasks. RadVLM already reports diverse automatic metrics -- RadGraph F1 and GREEN for report generation, macro-F1 for classification, IoU for grounding, and GPT-4o consistency for dialogue -- that can be used as individual or composite reward signals.  
We therefore envision fine-tuning RadVLM with multi-objective GRPO to align the model more tightly with clinical goals while mitigating potential reward-hacking artifacts through adversarial validation.  
A systematic exploration of this RL stage, including ablation on reward weighting and safety checks, is left to future work.

As radiology datasets grow in both scale and quality, and as vision-language modeling advances at a rapid pace, AI-driven tools are becoming ever more reliable and valuable for clinical practice. With RadVLM, we aim to contribute to this vibrant field by introducing a method that draws on prior work yet incorporates original ideas toward developing a multitask conversational vision-language model for radiology.

\section{Acknowledgments}
This work was supported as part of the Swiss AI Initiative by a grant from the Swiss National Supercomputing Centre (CSCS) under project ID a02 on Alps, and by the LOOP Zurich as part of the application driver project supporting the LOOP Zurich Biomedical Informatics Platform (BMIP). ND and FN received research support from the Digitalization Initiative of the Zurich Higher Education Institutions (DIZH)- Rapid Action Call - under TRUST-RAD project. CB received research support from the Promedica Foundation, Chur, CH.
TS is supported by the grant \#2021-911 of the Strategic Focal Area “Personalized Health and Related Technologies (PHRT)” of the ETH Domain (Swiss Federal Institutes of Technology). HM, MN and KF are supported by JSPS KAKENHI (Grant  Number: 23KK0148).
AR is supported by the StimuLoop grant \#1-007811-002 and the Vontobel Foundation. MV and SL are supported by the Swiss State Secretariat for Education, Research, and Innovation (SERI) under contract number MB22.00047. MK is supported by the UZH Global Strategy and Partnerships Funding Scheme and a Research Partnership Grant with China, Japan, South Korea and the ASEAN region (RPG 072023\_18).

\bibliographystyle{apalike} 
\bibliography{main.bib} 

\newpage
\appendix

\section{Appendix}

\begin{suppfigure}[ht]
    \centering
    \begin{tikzpicture}
    
        \node[draw, rounded corners, text width=0.8\textwidth, inner sep=10pt] (bubble) {
        \small
I will provide you with some visual information about a chest X-ray (that you cannot see) and ask you to create a 5-10 round Q\&A conversation between a user and an AI assistant. Please follow these guidelines:

\begin{enumerate}
\item You have exclusive access to certain details about a chest X-ray (\texttt{<INFORMATION\_ABOUT\_XRAY>}). These are the main findings (report), the view, and the gender of the Chest X-ray. The user does not have these details.
\item You will create a 5-10 round Q\&A conversation about the X-ray.
\item The user is a layperson and asks general, exploratory questions (e.g., what do you see, is it normal, etc.).
\item Your answers should be cautious, concise, and based only on what can be visually inferred from the provided details.
\item If coordinates for any visual findings are given, ensure they follow the provided format $[x_{min}, y_{min}, x_{max}, y_{max}]$. It should invite the assistant to answer about it, such as asking \textit{``Where is [observation] located on the image?''}
\item Do not offer definitive diagnoses or overconfident conclusions. Use careful, tentative language.
\item Do not reference any external data or medical reports beyond what is in \texttt{<INFORMATION\_ABOUT\_XRAY>}.
\item 8. **Do not reference or imply any prior studies or use the word “changes.”** Treat this as the only X-ray image available; (...)
\item Output strictly in this JSON format with pairs of messages:

\texttt{[
\{ "from": "human", "value": "Question text" \},
\{ "from": "gpt", "value": "Answer text" \},
...]
}

\item Do not include any additional text or comments. Only the JSON array of Q\&A pairs.
\end{enumerate}

\texttt{<INFORMATION\_ABOUT\_XRAY>}: 

\begin{itemize}
    \item \textbf{Radiology report:} Severe pneumonia left lung, and a smaller region of pneumonia (...)
    \item \textbf{List of Abnormalities:} Lung Opacity, (...)
    \item \textbf{View:} AP
    \item \textbf{Gender:} female
    \item \textbf{Selected observations with bounding boxes coordinates:} 
       Severe pneumonia left lung: [0.50, 0.48, 0.85, 0.81]
\end{itemize}
\textbf{Conversation in expected format:} 

        };
    \end{tikzpicture}

    \caption{\textbf{Prompting LLM to generate conversation data.}}
    \label{supp_fig:conv_prompt}
\end{suppfigure}

\begin{supptable}[t]
\centering
\footnotesize
\renewcommand{\arraystretch}{1.5} 
\begin{tabular}{|c|l|}
\hline
\textbf{Model} & \textbf{Link to Repository} \\ \hline
LLaVA-OV   & \url{https://huggingface.co/llava-hf/llava-onevision-qwen2-7b-si-hf} \\
LLaVA-Med  & \url{https://github.com/microsoft/LLaVA-Med} \\
RaDialog   & \url{https://huggingface.co/ChantalPellegrini/RaDialog-interactive-radiology-report-generation} \\
CheXagent  & \url{https://huggingface.co/StanfordAIMI/CheXagent-2-3b} \\
MAIRA-2    & \url{https://huggingface.co/microsoft/maira-2} \\ 
\textbf{RadVLM} & \url{https://huggingface.co/KrauthammerLab/RadVLM} \\ \hline
\end{tabular}
\caption{Link to repository for each model's weights}
\label{supptable:baseline-models}
\end{supptable}

\begin{suppfigure}[ht]
    \centering
    \begin{tikzpicture}
        \node[draw, rounded corners, text width=0.8\textwidth, inner sep=10pt] (bubble) {
            \small

You are an evaluator for a vision-language model. The model answers multi-round questions based on some reference data (referred to as “provided data”). For each round in the conversation, you have:

\begin{itemize}
	\item User's question
	\item Expected answer (if the model had access to the data)
	\item Generated answer (from the model that does not have direct access to the data, but only to the image)
\end{itemize}
Your task is to:
\begin{itemize}
	\item Explain your reasoning (very concisely) step by step (“chain of thought”) for each generated answer. This can include how you compare it to the expected answer, what details you notice, and how you decide on correctness, completeness, and relevance.
	\item Assign a small “per-question” rating or a short note indicating how close the generated answer is to the expected answer. Of course, we don't expect the generated answer to be perfect, so don't be too hard if there are small deviations.
	\item After analyzing all rounds, produce a single, final overall numeric score (an integer) from 0 to 10.
\end{itemize}

In your final output, provide:
\begin{itemize}
	\item Your detailed, step-by-step reasoning for each round.
	\item A single line at the end in the exact format:\\
\texttt{Overall score: <score>/10}
\end{itemize}
Include no extra lines or text beyond this.

\textbf{Provided Data:}
\begin{itemize}
    \item List of Abnormalities: Cardiomegaly, Lung Opacity, Pleural Other, Support Devices
    \item Radiology report: There is a large amount of air (...)
    \item View: AP
    \item Gender: female
    \item Selected observations with bounding boxes coordinates:
    \begin{itemize}
        \item Moderate right pneumothorax: [0.19, 0.15, 0.51, 0.41]
    \end{itemize}
\end{itemize}

\textbf{Conversation to Evaluate:}
\begin{itemize}
    \item \textcolor{blue}{User: Can you tell me if this chest X-ray looks normal?}
    
    \textcolor{red}{Expected answer: The X-ray has several findings that appear abnormal, including air (...).} 
    
    \textcolor{gray}{Generated answer: The X-ray shows some findings that are not typical, including (...)}
   
   \item \textcolor{blue}{User: What does air in the pleural space mean?}
   
   \textcolor{red}{Expected answer: Air in the pleural space typically indicates what's known (...)}
   
   \textcolor{gray}{Generated answer: Air in the pleural space, known as a pneumothorax, means (...)}

\end{itemize}

\textbf{Overall score:} [score]

        };
    \end{tikzpicture}

    \caption{\textbf{Prompting LLM to evaluate conversation inference of a VLM}}
    \label{supp_fig:eval_conv_prompt}
\end{suppfigure}

\begin{supptable}[h]
\centering
\footnotesize
\renewcommand{\arraystretch}{1.3}
\begin{tabular}{p{4cm}p{2cm}p{8cm}}
\toprule
\textbf{Task} & \textbf{Model} & \textbf{Prompt Used} \\ \midrule
\multirow{5}{*}{Report Generation}
 & LLaVA-OV & RadVLM’s prompts \\
 & LLaVA-Med & RadVLM’s prompts \\
 & RaDialog & “Write a radiology report for this X-Ray.” \\
 & CheXagent & “Write an example findings section for the CXR” \\
 & MAIRA-2 & HF template: processed inputs passed directly to model (no explicit prompt)  \\ \midrule
\multirow{2}{*}{Abnormality Classification}
 & CheXagent & “Identify any diseases visible in the given CXR. Options:\newline
   atelectasis, cardiomegaly, consolidation, edema, enlarged cardiomediastinum, fracture, lung lesion, lung opacity, pleural effusion, pleural other, pneumonia, pneumothorax, support devices” \\
 & RaDialog & “List all the findings in this report.”;\newline
   “Enumerate the observations from the report.”;\newline
   “What findings can be identified from this report?” \\ \midrule
\multirow{2}{*}{Abnormality Grounding}
 & CheXagent & Multiple templates, e.g.:\newline
   “Detect \{\} in the given image.”;\newline
   “Locate areas in the chest X-ray where \{\} is present, using bounding box coordinates.”;\newline
   “Localize \{\} in bounding box format for the given image.”;\newline
   “Find the locations of \{\} in the bounding box format for the given image.”;\newline
   “Locate \{\} for the given image.”;\newline
   “Examine the chest X-ray and mark the regions affected by \{\} with bounding boxes.”;\newline
   “Detect the following in the image: \{\}.”;\newline
   “Examine the image for regions affected by \{\}, and indicate their positions with bounding boxes.”;\newline
   “Perform detection for \{\}.”;\newline
   “Abnormality Grounding (VinDr-CXR): \{\}.” \\
 & MAIRA-2 & HF template: processed inputs passed directly to model \\ \midrule
\multirow{2}{*}{Anatomical / Phrase Grounding}
 & CheXagent & “Please locate the following anatomical region: \{\}”;\newline
   “Identify the position of the following region in the CXR: \{\}” \\
 & MAIRA-2 & HF template: processed inputs passed directly to model \\ \midrule
Conversations & All models & Generated questions from GPT-4o \\ \bottomrule
\end{tabular}
\caption{\textbf{Prompts used for each baseline model across tasks.} Summary of the input prompts (or templates) employed for inference by each compared model. These templates were obtained from released material (paper, GitHub repo, etc.) when available; otherwise, RadVLM's prompts were used (for LLaVA-OV and LLaVA-Med). “HF” denotes HuggingFace.}
\label{supptable:prompts-comparison}
\end{supptable}

\end{document}